\documentclass[a4paper,11pt,final]{article}

\usepackage{graphicx,amsmath,amssymb,algorithm,algorithmic,textcomp,tikz-cd,float} 
\usepackage{epsfig}

\begin{document}

\title{\ \\ \LARGE\bf Analyzing symmetry and symmetry breaking by computational aesthetic measures}

\author{Hendrik Richter \\
HTWK Leipzig University of Applied Sciences \\ Faculty of
 Engineering\\
        Postfach 301166, D--04251 Leipzig, Germany. \\ Email: 
hendrik.richter@htwk-leipzig.de }

\maketitle

\begin{abstract} 
We study creating and analyzing symmetry and broken symmetry in digital art. Our focus is not so much on computer--generating  artistic images, but rather on  analyzing  concepts  and templates for incorporating symmetry and symmetry breaking into the creation process. Taking as a starting point patterns generated algorithmically by emulating the collective feeding behavior of sand--bubbler crabs, all four types of two--dimensional symmetry are used as isometric maps. Apart from a geometric interpretation of symmetry, we also consider color as an object of symmetric transformations. Color symmetry is realized as a color permutation consistent with the isometries. Moreover, we analyze the abilities of computational aesthetic measures to serve as a metric that reflects design parameters, i.e. the type of symmetry and the degree of symmetry breaking.

\end{abstract}

\section{Introduction}

It is generally acknowledged that symmetry is an important property of aesthetically pleasing objects~\cite{al17,gruen86,schatt17,weil52}. Thus, if we intend to generate objects of vi\-sual art and employ a computational  measure to quantify their aesthetic value, then the  measure should be able to capture conceptually and computationally different aspects and meanings of symmetry. This paper considers digital art with pre--designed and tuneable degrees of symmetry and studies the abilities of  computational   aesthetic measures to identify such different shades of symmetry. 

There are several issues with both computationally evaluating and  algorithmically  generating  symmetry in  digital visual art. 
Symmetry implies that between point sets forming the geometric objects of an image there are isometric maps preserving certain properties of the point sets. In other words, parts of the image resemble each other  in some way or another. Thus, the main procedure to search for symmetry is to find similarity (or even equality) between parts of the image by comparing them. This can be done by dividing the image into sections, for instance an upper and a lower section, or a left and a right, or into sections partitioned by the diagonals. Subsequently, we may use a measure to evaluate the sections pixel--wise, for instance by the intensity value of the pixels or their luminosity. We then compare the values obtained for each section. The more similar the values of the measure are, the more substantial, it is supposed, is the symmetry.  The approach can be refined by comparing not only two sections, but several more. However, the main challenge  is that such an evaluation of symmetry in some ways prejudges our detection abilities as we preset the subsections supposed to appear particularly similar and thus symmetric. Recently, an alternative method has been proposed that uses a swarm--based search algorithm to find promising focal points of the image which are optimal in showing symmetry~\cite{al17}.   Consequently, we no longer need to preselect sections of the image to be tested for symmetry.  However, also this method assumes symmetry with invariant points such as rotation with a rotation center point or reflection with a reflection axis. In addition, it is computationally more expensive. 

Although symmetry of two--dimensional geometric objects can be easily defined mathematically by isometric maps,  generating interesting symmetry is not trivial. An art object that strictly adheres to the mathematical definition of symmetry  frequently appears to be not overly aesthetic from a human point of perception.  Images with perfect symmetry sometimes seem ``overdesigned''. Another related  problem is that symmetry in a strict mathematical sense is a binary concept. Either there is symmetry and  the objects in an image obey an isometric map, or there is not. However, human artists creating works  that are praised for handling subtle effects of symmetry often experiment with symmetries that are slightly (or even  substantially) perturbed~\cite{ada16,bier05,mol86,schatt04}. Such perturbations  can be seen as   
symmetry breaking. From a rather abstract point of view, 
symmetry breaking is not meaning that symmetry is completely absent or that there is asymmetry, but rather that some aspects of symmetry are gone. Symmetry breaking is a particularly powerful concept if seen as a process that plays with our expectations of symmetry and thus needs its context. Put differently, if before symmetry breaking there was one kind of symmetry, then a broken symmetry implies another kind  of, but somehow ``lesser'' symmetry.  

Recently, an algorithmic framework has been proposed to generate visual art which is based on the collective feeding behavior of sand--bubbler crabs~\cite{rich18}. In nature, these patterns consist of sand--balls. In the images inspired by nature, the patterns consist of pellets with a given color (and possibly texture). This paper deals with employing digital sand--bubbler patterns for studying algorithmic generation and computational evaluation of symmetry and symmetry breaking. We may generate symmetry by applying to the patterns any of the four types   of isometric symmetry  in two--dimensional objects (e.g.~\cite{liu10}): (i) reflection, (ii) rotation, (iii) translation, and (iv) glide reflection. To have images with broken symmetry we  remove (or render invisible) a fraction of the pellets building the patterns. 
This interpretation is based on ideas proposed by Molnar \& Molnar~\cite{mol86} in the 1980s that symmetry breaking in visual objects can be realized by moving or removing building blocks of the visual representation. A main advantage of such an interpretation is that by fixing a fraction of pellets and removing the amount of pellets thus allowed at random, the degree of symmetry breaking can be almost continuously scaled. If the fraction is zero, symmetry is completely intact, any fraction between zero and one breaks the symmetry to that degree, and if the fraction is equal to one, symmetry is entirely absent. We will use such a scaling to  pre--design and tune degrees of symmetry in sand--bubbler patterns.  

As the sand--bubbler patterns have a given color, it appears interesting to consider also color as a property that may undergo symmetry transformations. This is known as color symmetry~\cite{schwarz84,sen83,sen88}.
There are some works on creating patterns using color symmetry, for instance~\cite{dun10,ou12,thom12}, and a substantial amount of the visual art of M.C. Escher uses color symmetry in some way or another~\cite{ada16,cox86,schatt04,schatt17}, but there is a (somehow surprising) lack of applications in the domain of generative and evolutionary art.  In visual art color symmetry is a permutation of the patterns' colors which is consistent with the symmetry of the geometric objects of the image. It can be considered as a mapping on a color wheel. Color symmetry breaking, in turn, is an (intentional or random) perturbation of the color permutation. 
The visual and numerical results reported in this paper also intend to experiment with color symmetry in generative visual art. 

The paper is organized as follows. In Sec. \ref{sec:bubbler} the generation of sand--bubbler pattern is briefly recalled, see~\cite{rich18} for details. It is also discussed how generating and breaking symmetry can be achieved for these patterns. Computational experiments and results are presented in Sec. \ref{sec:results}. The paper is concluded with a summary of the findings and a discussion about future work.

\section{Generating symmetry and symmetry breaking in sand--bubbler patterns} \label{sec:bubbler}
\subsection{Pattern symmetry}

Sand--bubbler are tiny crabs living on tropical beaches. They create remarkable patterns in the sand as part of their collective feeding behavior.  
According to, and adopting, the language of biological field work, these patterns consist of sand balls, called pellets, that are placed along lines, called trenches, which radiate from a center point, called burrow. 
Recently, it was proposed to let this behavior inspire an algorithmic framework for generating visual art~\cite{rich18}. In this paper we use this framework for experimenting with symmetry and symmetry breaking, and  briefly recall  how the patterns are generated.

A sand--bubbler pattern can be described  by the pellets it contains. We give every pellet  a location $(x_{ijk},y_{ijk})^T$ in a two--dimensional plane. The index $(ijk)$ identifies the $k$--th pellet ($k=1,2,3,\ldots,K_{ij}$) belonging to the $j$--th trench ($j=1,2,3,\ldots,J_i$) of the $i$--th burrow  ($i=1,2,3,\ldots,I$). A pellet location can be computed by
\begin{equation} \left( \begin{array}{c} x_{ijk} \\ y_{ijk }\end{array}\right)=  \left( \begin{array}{c} x_i+r_k \cdot \cos{(\theta_j)} \\ y_i+ r_k \cdot \sin{(\theta_j)} \end{array}\right) +  \left( \begin{array}{c} \mathcal{N}(\mu_{ijk},\sigma_{ijk}^2) \\  \mathcal{N}(\mu_{ijk},\sigma_{ijk}^2)  \end{array}\right) , \label{eq:pellet}\end{equation}
where $(x_i,y_i)^T$ are the coordinates of the $i$--th burrow, $\theta_j$ is the trench angle of the $j$--th trench, $r_k$ is the radial coordinate of the $k$--th pellet, and  $\mathcal{N}(\mu_{ijk},\sigma_{ijk}^2)$ are realizations of a  random variable normally distributed with mean $\mu_{ijk}$ and variance $\sigma_{ijk}^2$, see~\cite{rich18} for details. For each burrow, we need to specify the maximum number of pellets $K_{ij}$ for a given $i$ and $j$; the same applies to the maximum number of trenches $J_i$.

For two--dimensional objects represented in an Euclidean space, there are four types of isometric symmetry (e.g.~\cite{liu10}): (i) reflection, (ii) rotation, (iii) translation, and (iv) glide reflection. For a sand--bubbler pattern specified by Eq. (\ref{eq:pellet}), we define   $n$ an integer and $(\Delta x,\Delta y)$ some real numbers,  compute $\varrho=\sqrt{x_{ijk}^2+y_{ijk }^2}$,  and  obtain 
(left--right) 
reflection,  rotation, translation, and (up--down) glide reflection by \begin{align} (x_{ijk}, y_{ijk })^T &\rightarrow (-x_{ijk}, y_{ijk })^T  \label{eq:reflect} \\
(x_{ijk}, y_{ijk })^T &\rightarrow (\varrho \cos{(2\pi/n)}, \varrho \sin{(2\pi/n)})^T \label{eq:rotate} \\
 (x_{ijk}, y_{ijk })^T &\rightarrow (x_{ijk}+\Delta x, y_{ijk }+\Delta y)^T \label{eq:transl} \\
 (x_{ijk}, y_{ijk })^T &\rightarrow (x_{ijk}+\Delta x, -y_{ijk })^T, \label{eq:glide} \end{align} respectively.
Note that for translation (\ref{eq:transl}) and glide reflection (\ref{eq:glide}), no point of the pattern remains invariant, while for reflection  (\ref{eq:reflect}) and rotation (\ref{eq:rotate}), there are invariant points with the reflection axis and the rotation center point, respectively. 
Basically, the definitions (\ref{eq:reflect})--(\ref{eq:glide}) of two--dimensional isometric symmetry relate to points $(x_{ijk}, y_{ijk })^T$ representing pellets, but may also apply  to point sets representing trenches, burrows or whole patterns. 
In this paper we consider symmetry only to act on whole burrows. 

As important as symmetry is for aesthetically pleasing objects, it is also known that real beauty in nature and art is sometimes connected with symmetry that is a little less than completely perfect. Look at beauty found in nature where symmetry surely is a major organizational principle, but is rarely achieved in a strict mathematical sense. The same applies to art, see for instance the discussion about oriental carpets, embroideries, tilings, and ornaments~\cite{bier01,bier05,gruen86}. Such slight imperfections of symmetry fall into an intermediate state between complete symmetry and absence from any symmetry and can be related to symmetry breaking.  Molnar \& Molnar~\cite{mol86} suggested that in visual art symmetry breaking can be achieved by moving and/or removing building blocks of the visual representation. This is in line with the observation that symmetry breaking in textile art (for instance oriental carpets and embroideries) may be created by intentionally or randomly inserting irregularities and perturbations, resulting in a (more or less close) approximation of symmetry~\cite{bier01,bier05}.

From a computational point of view, and applied to the sand--bubbler patterns as defined by Eq. (\ref{eq:pellet}), this interpretation of symmetry breaking has an interesting property. By fixing a fraction of pellets and removing these pellets, the degree of symmetry breaking can be almost continuously scaled. Thus, symmetry and symmetry breaking can be pre--designed and tuned, which is opening up settings for computational experiments. For generating images and evaluating them using computational aesthetic measures we employ and analyze a parameter governing symmetry and symmetry breaking: the symmetry breaking rate $\sigma_{break}$. It describes how many of the pellets of a pattern are removed (or not visible) due to symmetry breaking.   If $\sigma_{break}=0$, symmetry is completely intact, any value $0<\sigma_{break}<1$ breaks the symmetry to that degree, and if $\sigma_{break}=1$, symmetry is entirely absent. As there are in total $I_\Sigma=\sum _{i=1}^{I} \sum_{j=1}^{J_i} K_{ij}$  pellets belonging to the pattern, there are $\lceil  \sigma_{break} I_\Sigma \rceil$ pellets  taken away (or rendered invisible).

\subsection{Color symmetry}
Using the mathematical concept of algebraic groups and understanding symmetry as a mapping that preserves certain structures, the notion of symmetry can be expanded beyond strictly meaning point sets, for instance towards color symmetry or dilation. For the sand--bubbler patterns specified by Eq. (\ref{eq:pellet}),  color symmetry appears to be particularly interesting. This is because patterns as found in nature on tropical beaches are monochromatic. They have the color of the sand they are built from.   In the artistic interpretation recently suggested~\cite{rich18}, it was proposed to color the pellets according to the chronological order of the placement, or to give each burrow a specific color, or to apply another coloring scheme thinkable. In fact, color is not an intrinsic property of a pattern, but requires a design of its own.  
Thus, a colored sand--bubbler pattern needs to specify the color $c_{ijk}$ of each pellet location $(x_{ijk}, y_{ijk })^T$. The color may vary over pellets, or trenches, or burrows, or not at all.

Color symmetry~\cite{schwarz84,sen83,sen88} of a pattern means that the coloring of the geometric objects building the pattern is consistent with the symmetry properties of these objects. Suppose there is a symmetry group of a pattern, for instance the isometric symmetries $\phi$ acting on a pellet according to  $(x_{ijk}, y_{ijk })^T \rightarrow \phi (x_{ijk}, y_{ijk })^T$ as defined by Eqs. (\ref{eq:reflect})--(\ref{eq:glide}). Further assume the pellet has the color $c_{ijk}$. Then color symmetry implies that every $\phi$ is associated with a color permutation giving the symmetric pellet the color $\theta(c_{ijk})$. The mappings  $\phi$ and  $\theta$ are to be homomorphic.  Put differently, the symmetry properties of the geometric objects consistently define the coloring of these objects. The colors can be defined by a permutation group of colors. The color permutation can be realized by a mapping on a color wheel. Based on this understanding, breaking color symmetry of a pattern can be achieved similarly to the symmetry breaking of the geometric aspects of patterns as described above. We again set the number of pellets for which color symmetry applies by $\sigma_{break}$. A broken color symmetry means that for these pellets the color is not determined by the color permutation of the ``unbroken'' pellets, but is either produced by a different permutation or defined otherwise.  
The computational  experiments reported next also deal with color symmetry. 
To distinguish between pattern symmetry breaking and color symmetry breaking, we call $\sigma_{break}(p)$ the pattern symmetry breaking rate and $\sigma_{break}(c)$ the color symmetry breaking rate.

\begin{figure}[tb]
\center
\includegraphics[trim = 30mm 100mm 30mm 90mm,clip,width=4cm, height=4cm]{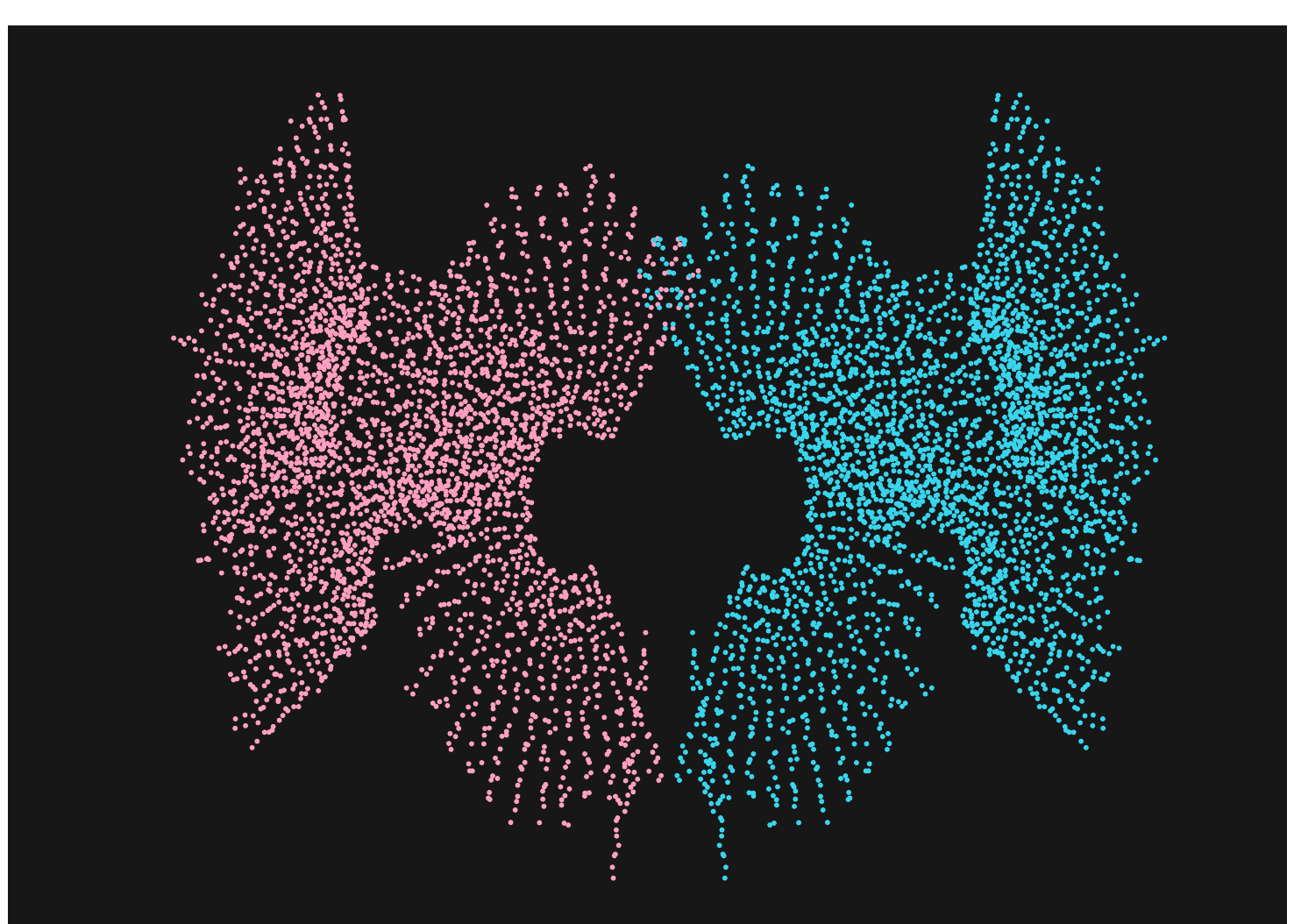}
\includegraphics[trim = 30mm 100mm 30mm 90mm,clip,width=4cm, height=4cm]{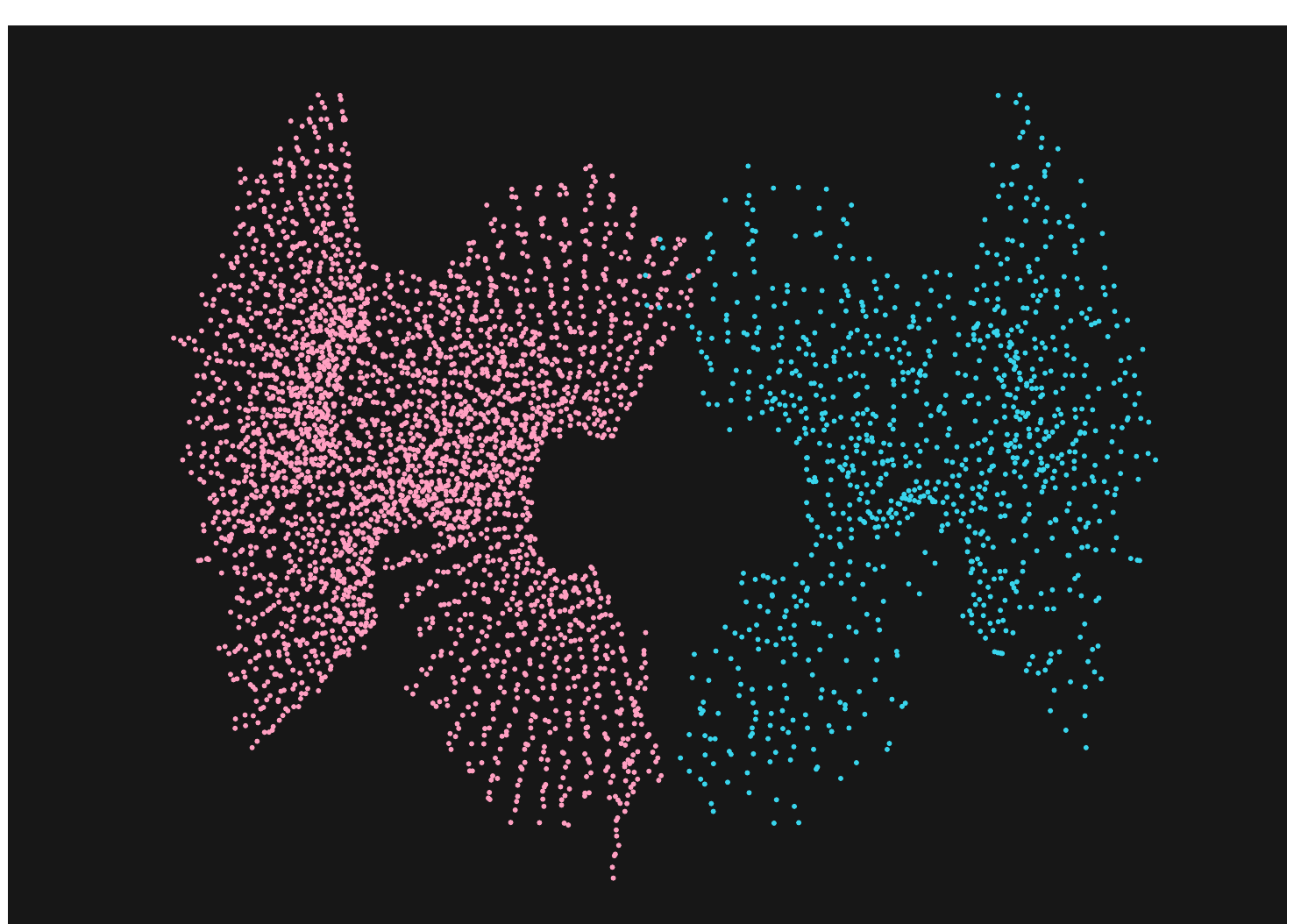}
\includegraphics[trim = 30mm 100mm 30mm 90mm,clip,width=4cm, height=4cm]{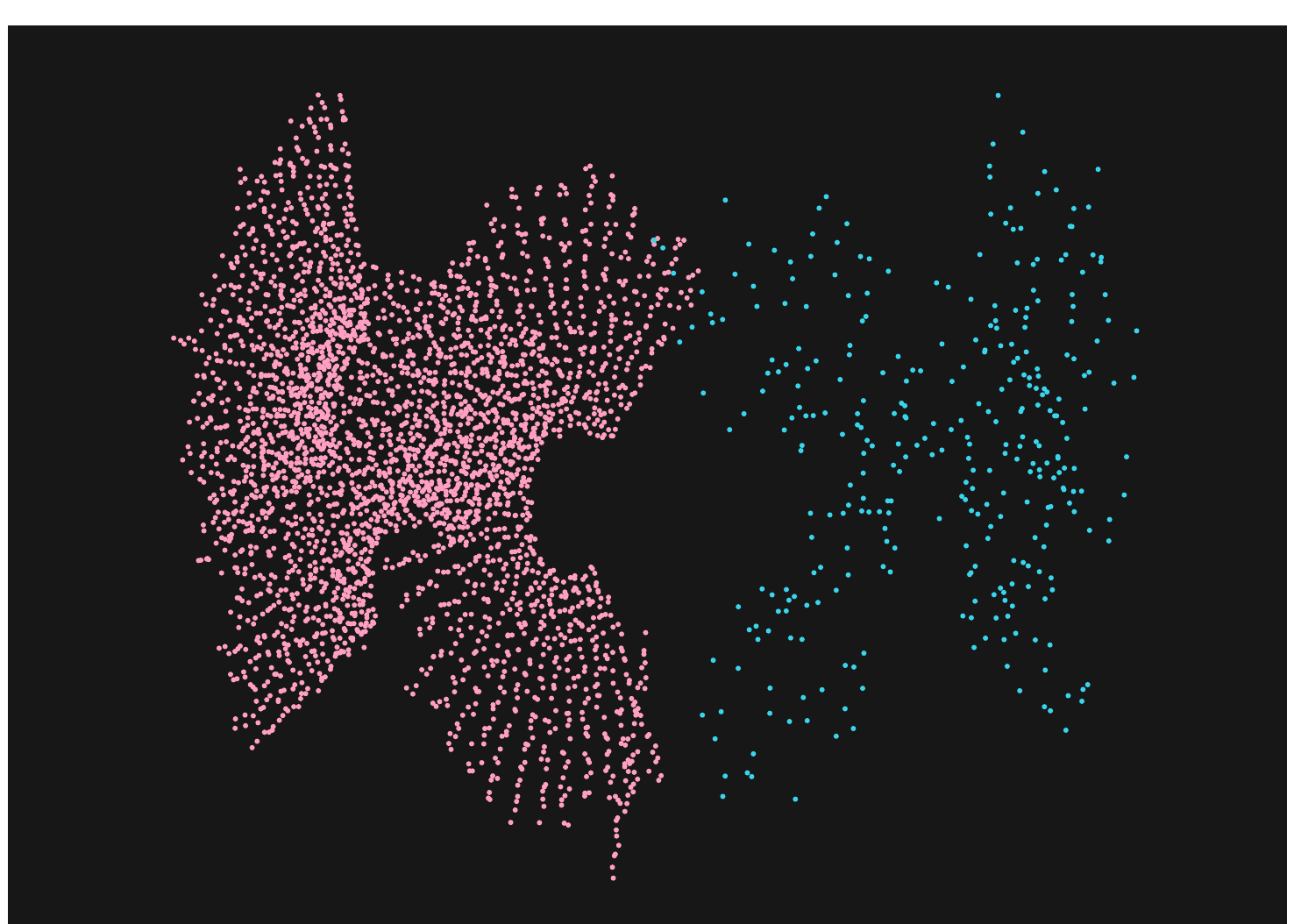}

 \hspace{0.1cm} (a) $\sigma_{break}(p)=0$ \hspace{1.3cm} (b) $\sigma_{break}(p)=0.1$ \hspace{1.25cm} (c) $\sigma_{break}(p)=0.3$
 
\includegraphics[trim = 30mm 100mm 30mm 90mm,clip,width=4cm, height=4cm]{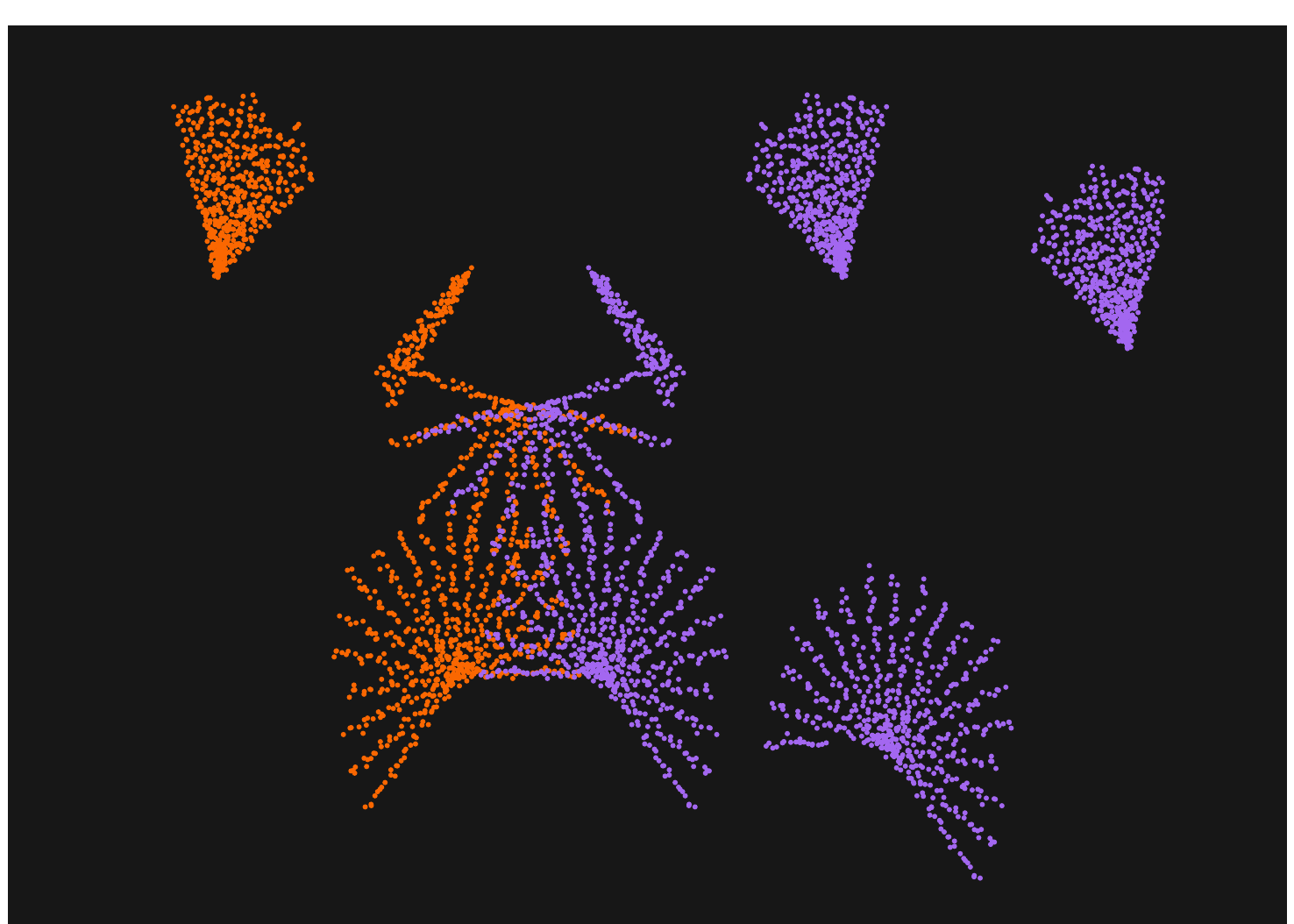}
\includegraphics[trim = 30mm 100mm 30mm 90mm,clip,width=4cm, height=4cm]{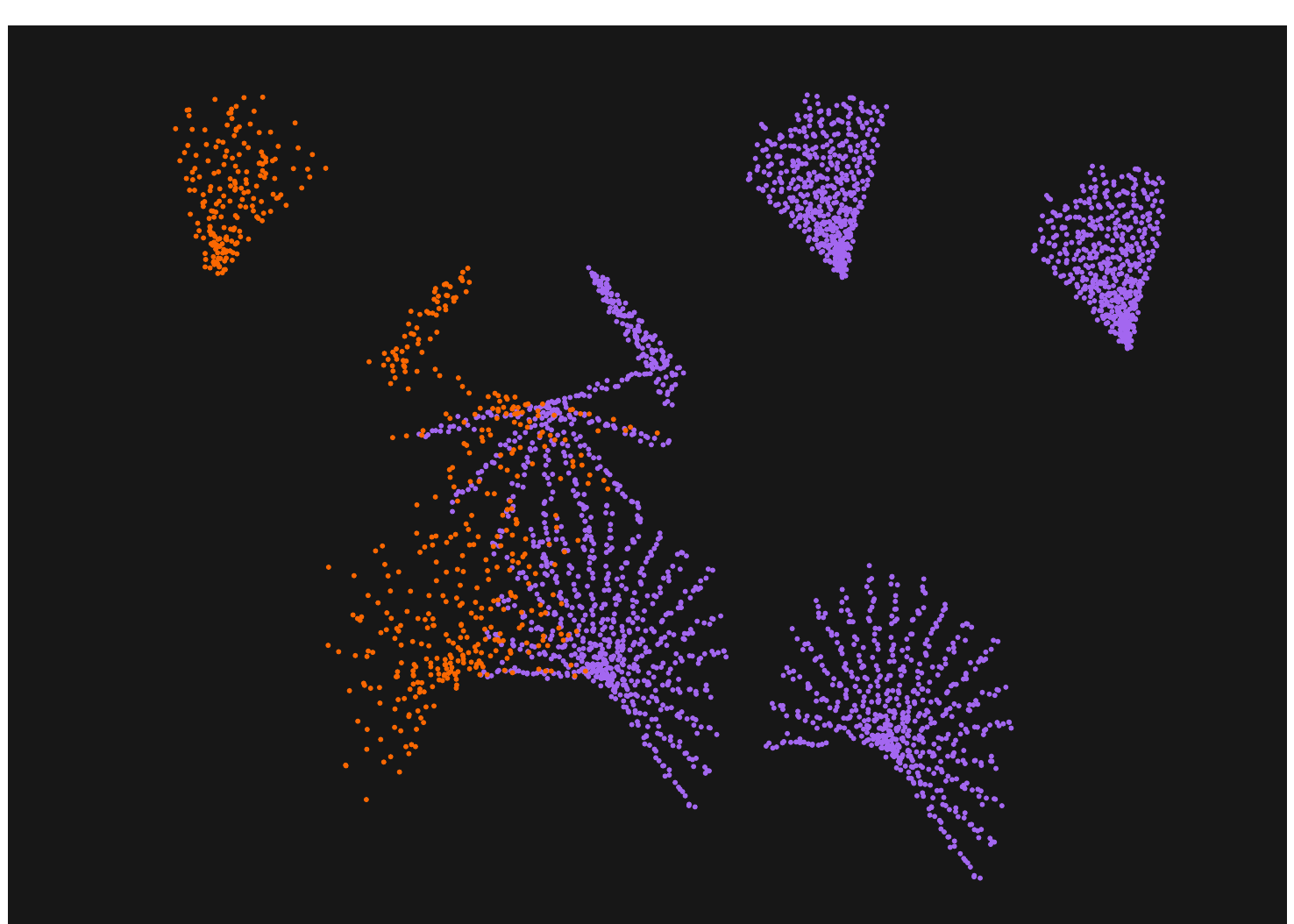}
\includegraphics[trim = 30mm 100mm 30mm 90mm,clip,width=4cm, height=4cm]{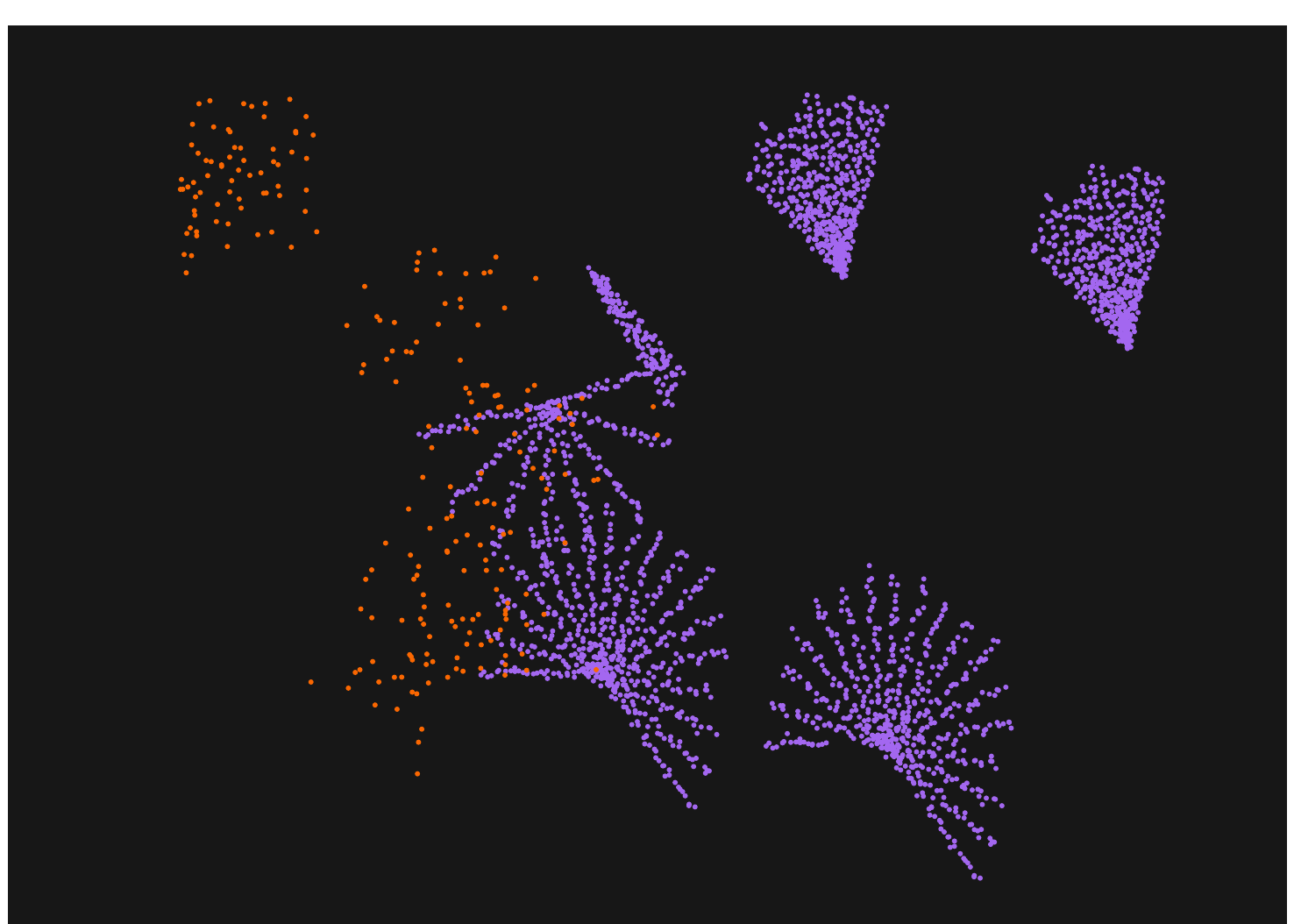}

 \hspace{0.1cm} (d) $\sigma_{break}(p)=0$ \hspace{1.3cm} (e) $\sigma_{break}(p)=0.1$ \hspace{1.25cm} (f) $\sigma_{break}(p)=0.3$

\caption{Images with dichromatic symmetry   
}
\label{fig:symm1}
\end{figure}

 \begin{figure}[tb]
 
 \center
\includegraphics[trim = 10mm 100mm 10mm 80mm,clip,width=5cm, height=4cm]{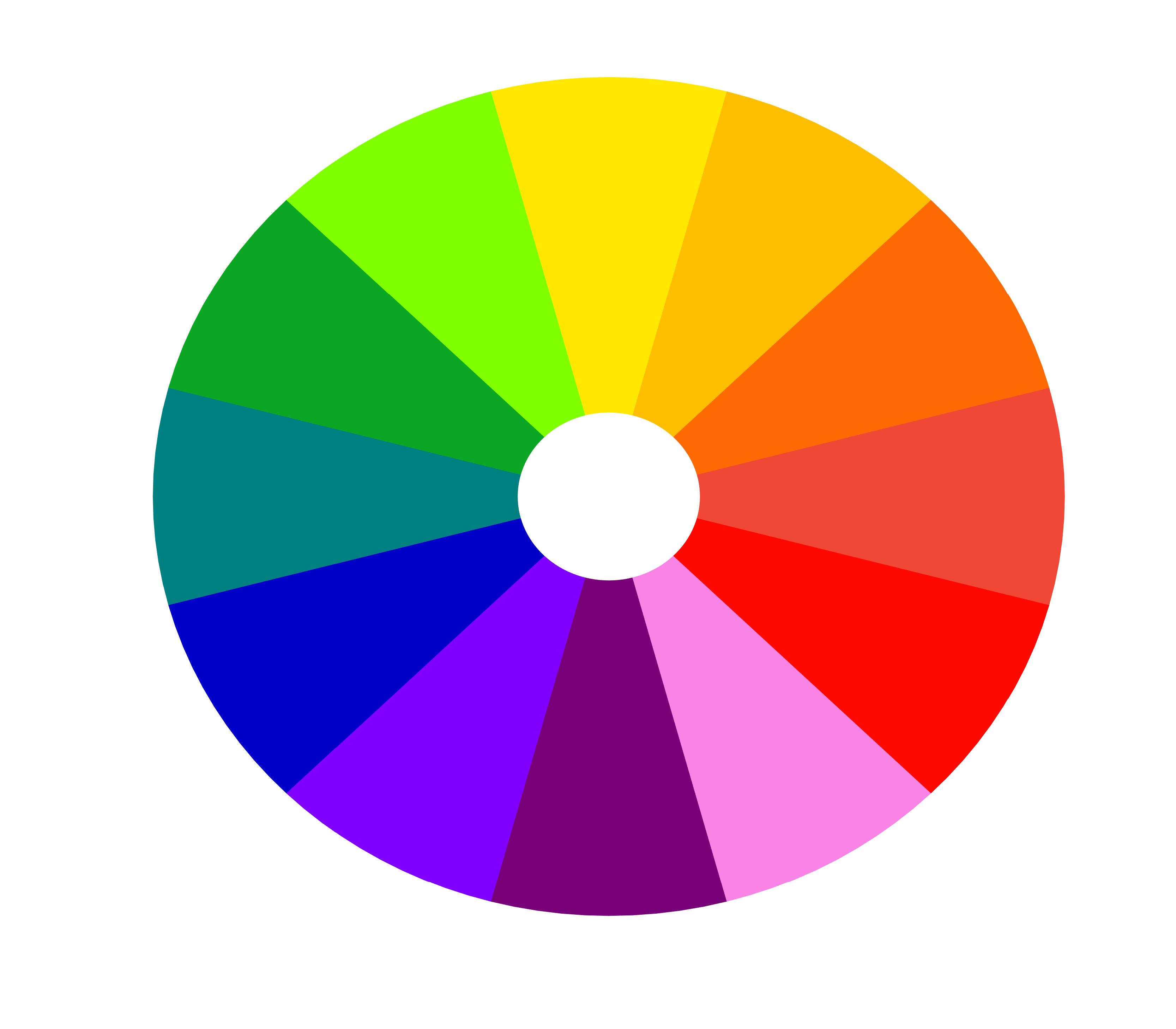}
\includegraphics[trim = 10mm 100mm 10mm 80mm,clip,width=5cm, height=4cm]{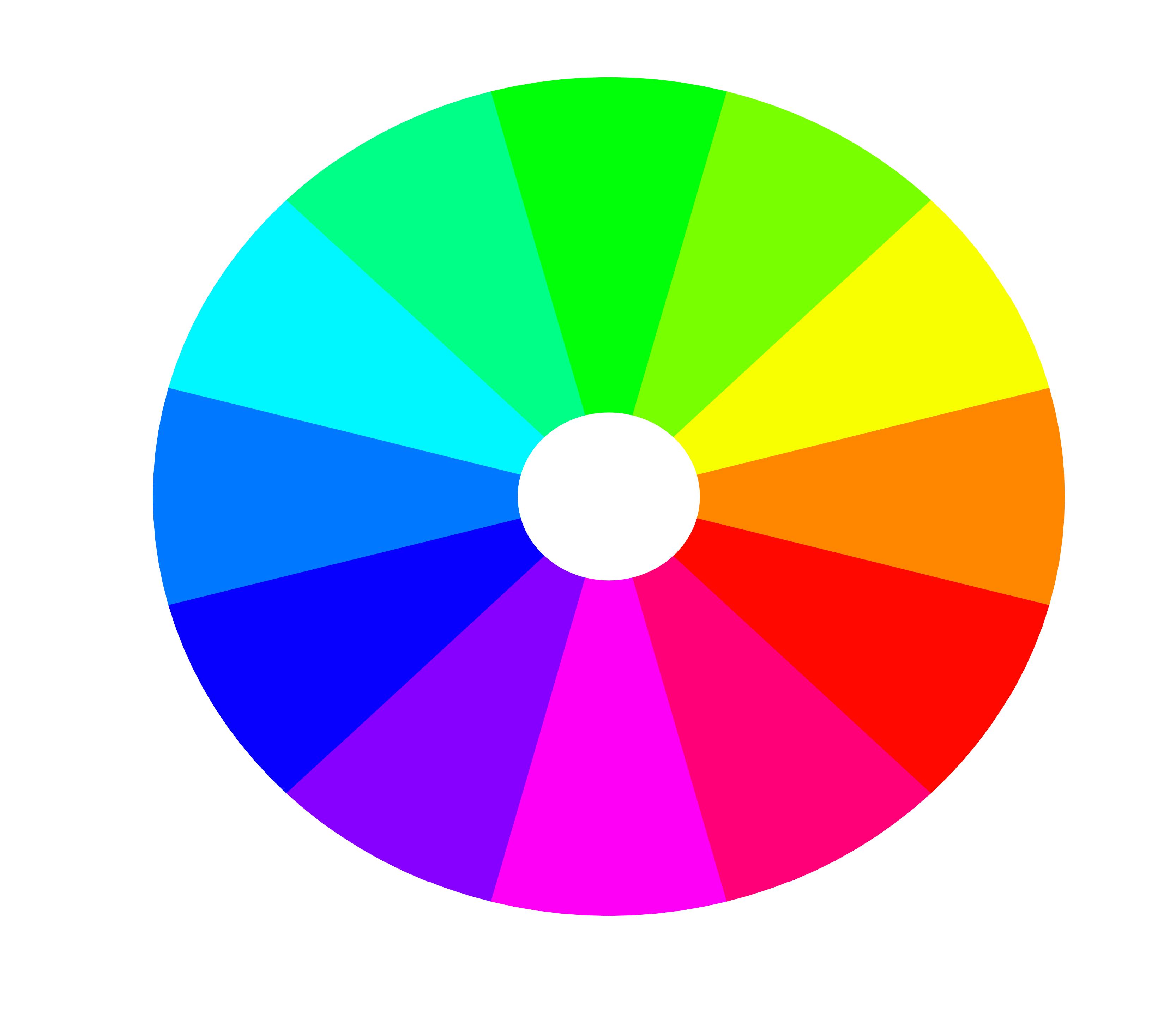}

(a) RYB \hspace{3.8cm} (b) RGB

\caption{Color wheels   with 12 slots, which are used to build color permutation groups with degree up to 12
}
\label{fig:col_wheel}
\end{figure}

We close the discussion with mentioning that there is a substantial body of work in physics, particularly particle physics, about symmetry, symmetry breaking and even color symmetry. This is because conservation of physical properties can be described as a consequence of symmetry in equations describing physical reality, and some special properties of some special particles were even ``christened colors, although of course they have nothing directly to do with color in the usual sense.''~\cite{wil15}, p.~232. This research  has no direct impact on the discussion in this paper. There is, however, an example where the spontaneous chiral symmetry breaking of electron emissions in polarized Cobalt was used as a template  to visualize a broken color symmetry~\cite{farsi07}.

\section{Computational experiments and results} \label{sec:results}

\subsection{Creating images with symmetry and broken symmetry}
For illustrating the effect of the different symmetries and symmetry breaking schemes discussed in Sec. \ref{sec:bubbler}, we consider some examples. We start with pattern symmetry, see Fig. \ref{fig:symm1} showing dichromatic images.  In principle, and if the background has a different color as the pellets and is not counted, displaying symmetry could also be possible in monochromatic images, but from an artistic as well as a computational  point of view such images might  not be very interesting.  Thus,  to begin with, our focus is on dichromatic symmetry.   The upper and lower left images (Fig. \ref{fig:symm1}a and  \ref{fig:symm1}d) show full symmetry by reflection, while in the lower image  there is additionally a translation. The middle and right panels  depict the same image with different degrees of symmetry breaking. In the upper panels (Fig. \ref{fig:symm1}b and  \ref{fig:symm1}c)  the symmetry breaking is achieved by removing (or making invisible) pellets, while in the lower panels (Fig. \ref{fig:symm1}e and  \ref{fig:symm1}f) only the reflected pellets undergo symmetry breaking and are additionally moved by realizations of a random variable. We see that the symmetry by reflection fades for the symmetry breaking rate getting larger, up to the point where it cannot be recognized anymore.

\begin{figure}[tb]
\center
\includegraphics[trim = 30mm 100mm 30mm 90mm,clip,width=4cm, height=4cm]{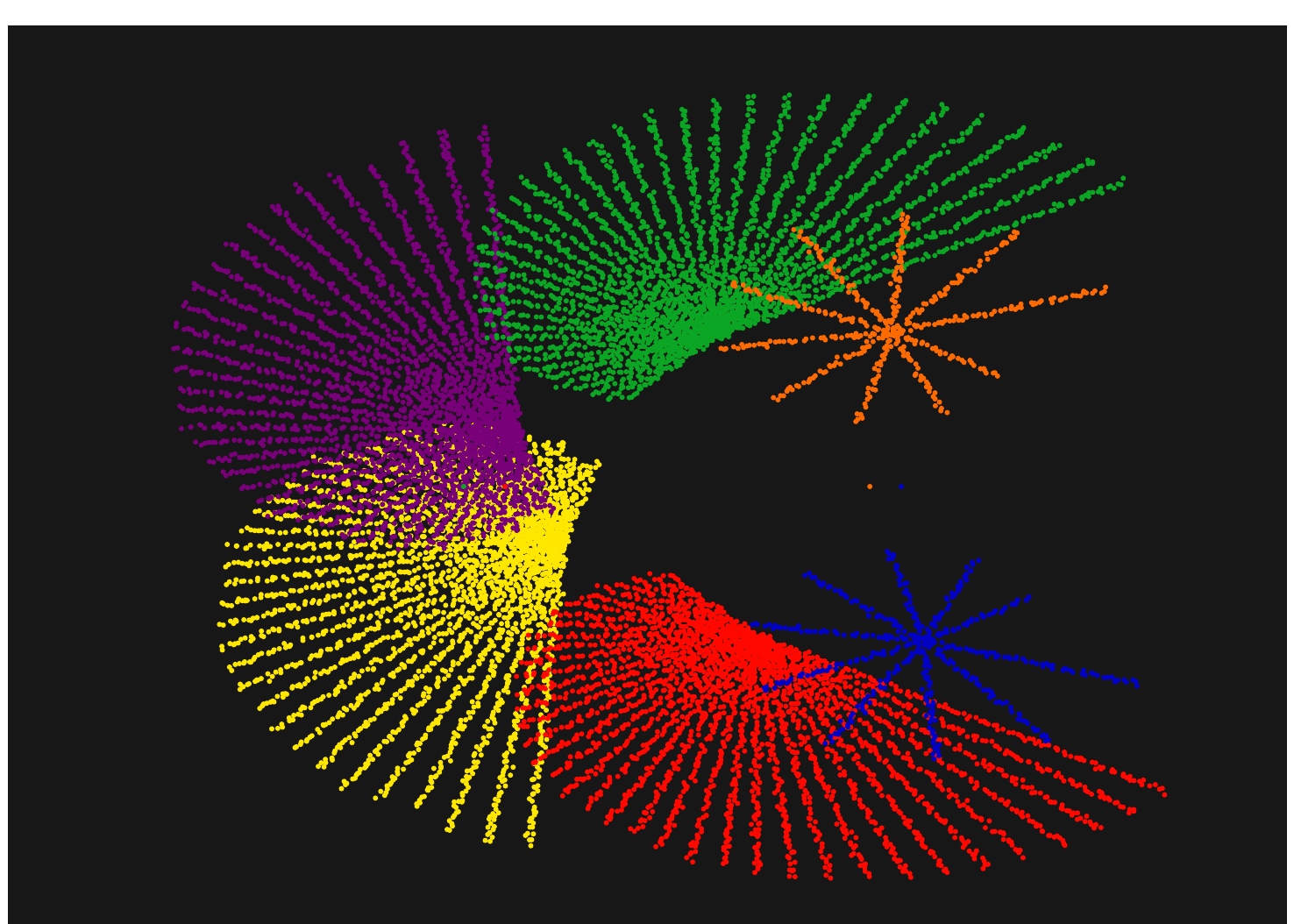}
\includegraphics[trim = 30mm 100mm 30mm 90mm,clip,width=4cm, height=4cm]{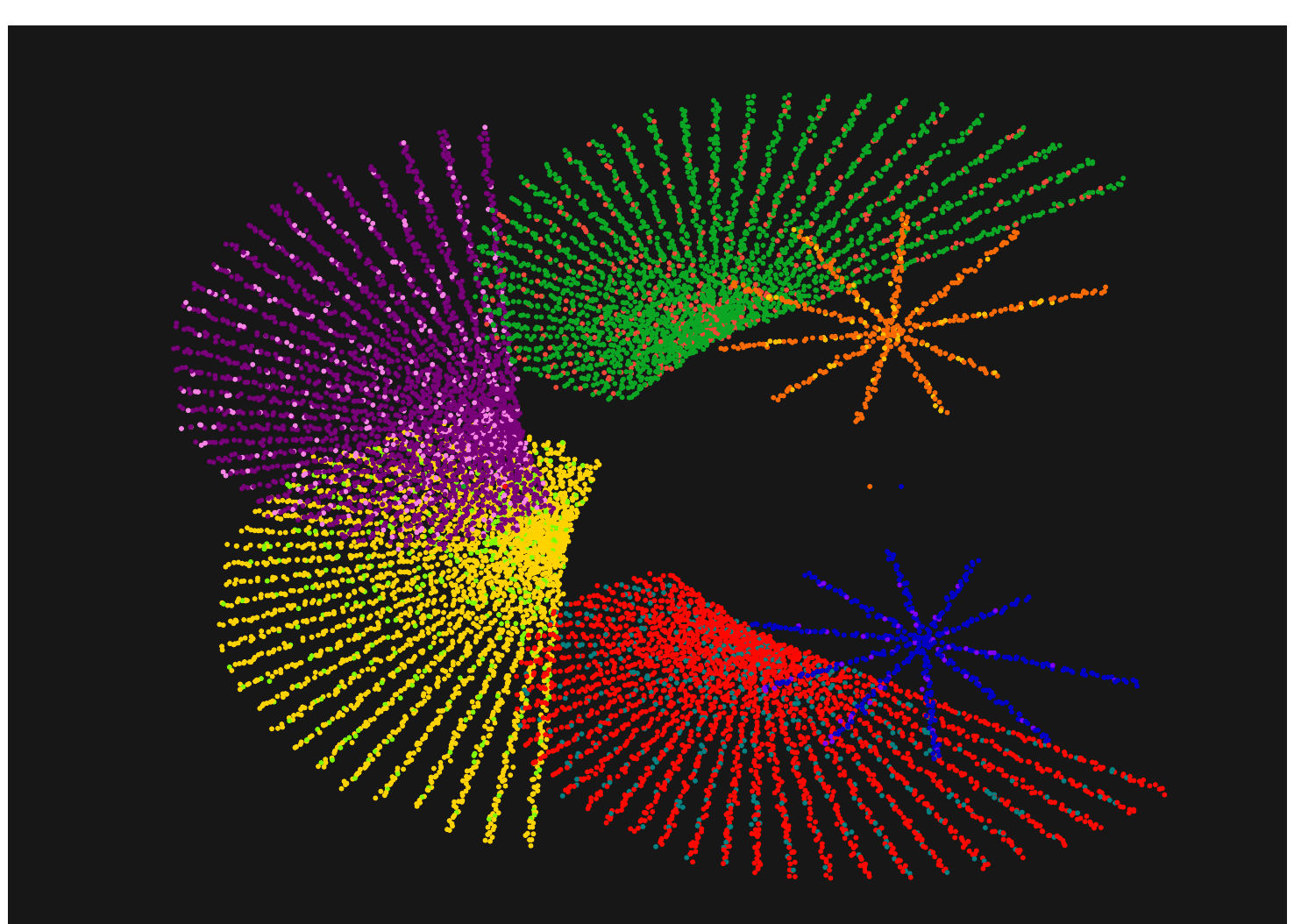}
\includegraphics[trim = 30mm 100mm 30mm 90mm,clip,width=4cm, height=4cm]{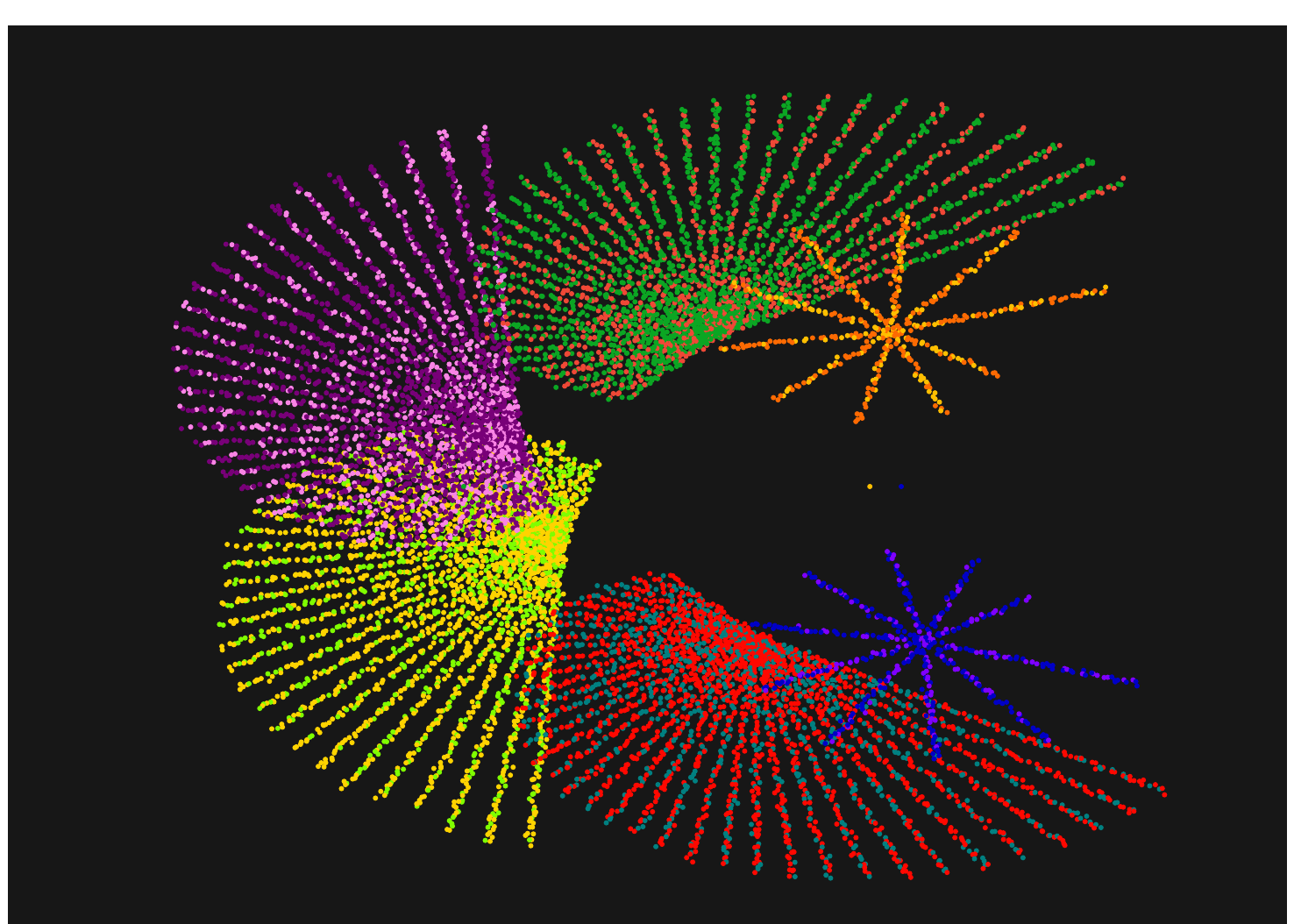}

 \hspace{0.1cm} (a) $\sigma_{break}(c)=0$ \hspace{1.7cm} (b) $\sigma_{break}(c)=0.15$ \hspace{1.25cm} (c) $\sigma_{break}(c)=0.35$

\includegraphics[trim = 30mm 100mm 30mm 90mm,clip,width=4cm, height=4cm]{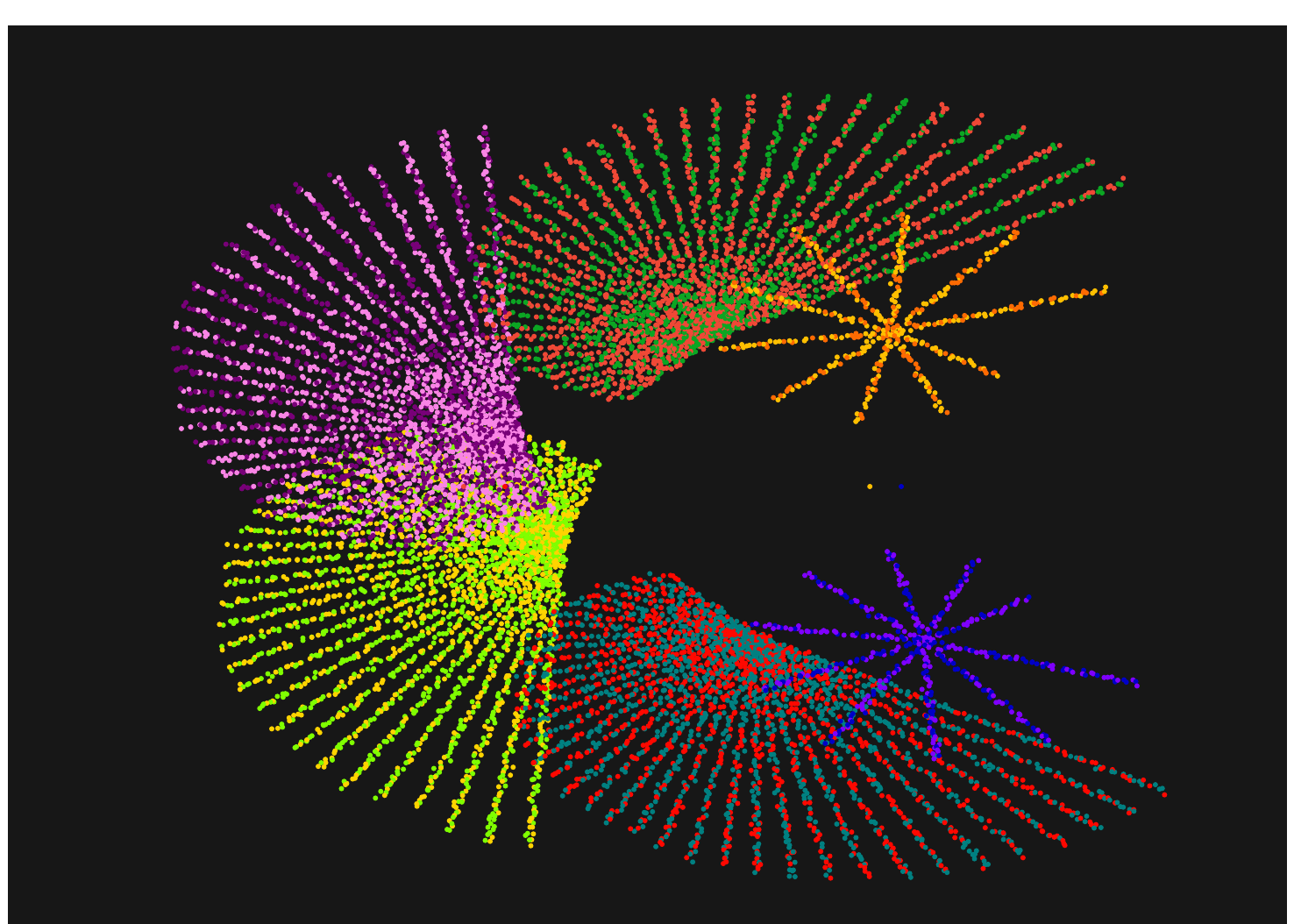}
\includegraphics[trim = 30mm 100mm 30mm 90mm,clip,width=4cm, height=4cm]{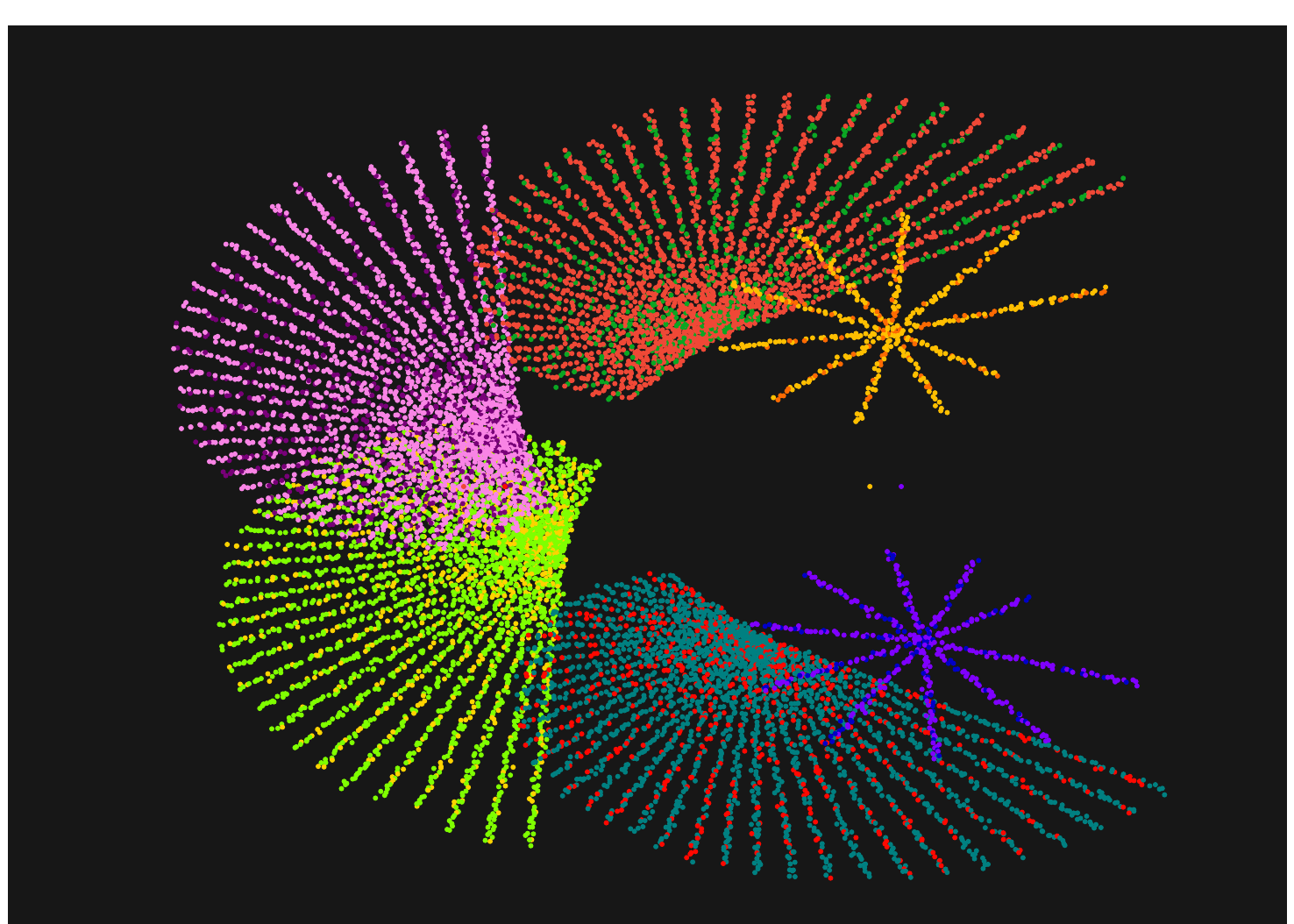}
\includegraphics[trim = 30mm 100mm 30mm 90mm,clip,width=4cm, height=4cm]{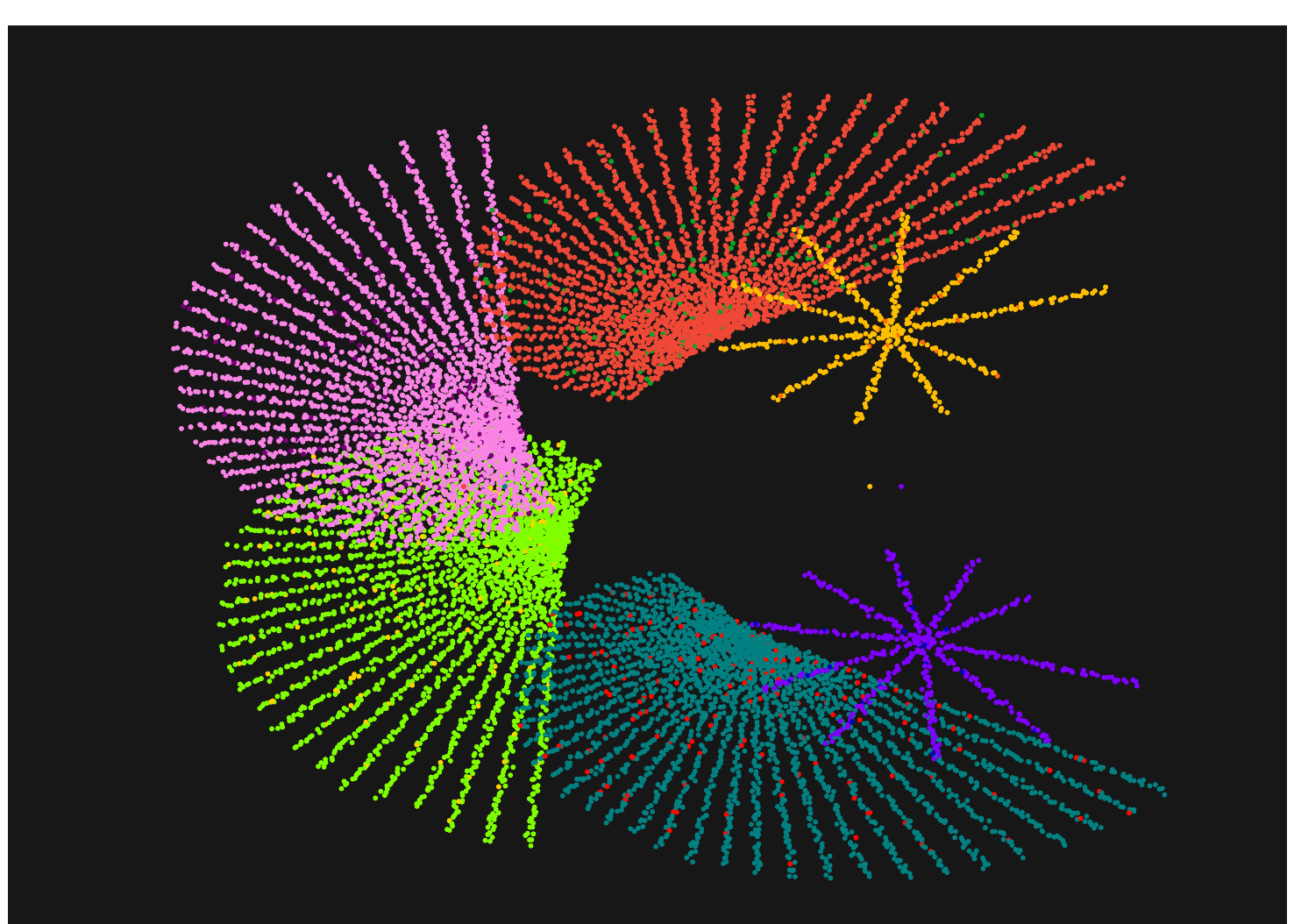}

 \hspace{0.1cm} (d) $\sigma_{break}(c)=0.55$ \hspace{1.25cm} (f) $\sigma_{break}(c)=0.75$ \hspace{1.25cm} (f) $\sigma_{break}(c)=0.95$

\caption{Images with polychromatic symmetry using a RYB color wheel   
}
\label{fig:symm2}
\end{figure}

\begin{figure}[tb]
\center
\includegraphics[trim = 50mm 100mm 50mm 90mm,clip,width=4cm, height=4cm]{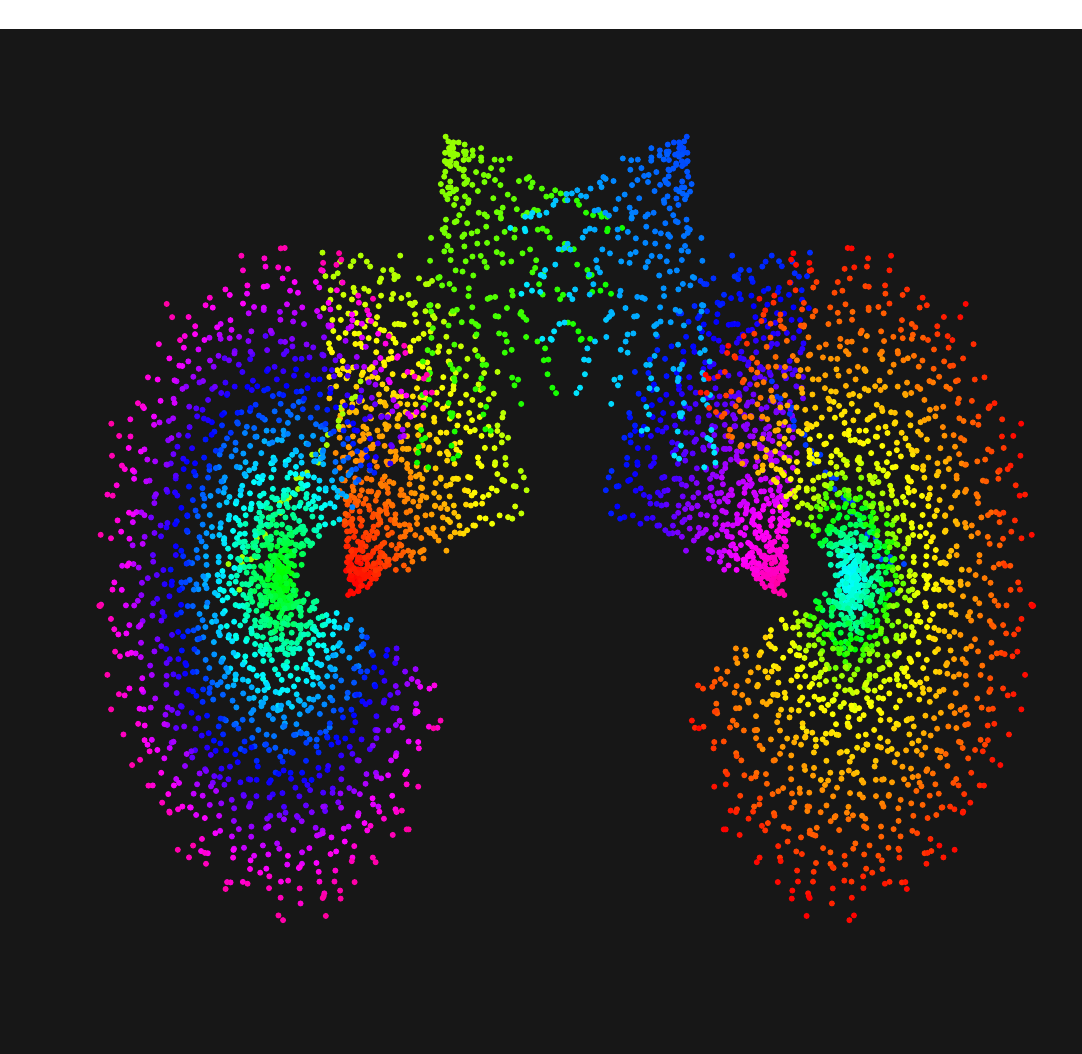}
\includegraphics[trim = 50mm 100mm 50mm 90mm,clip,width=4cm, height=4cm]{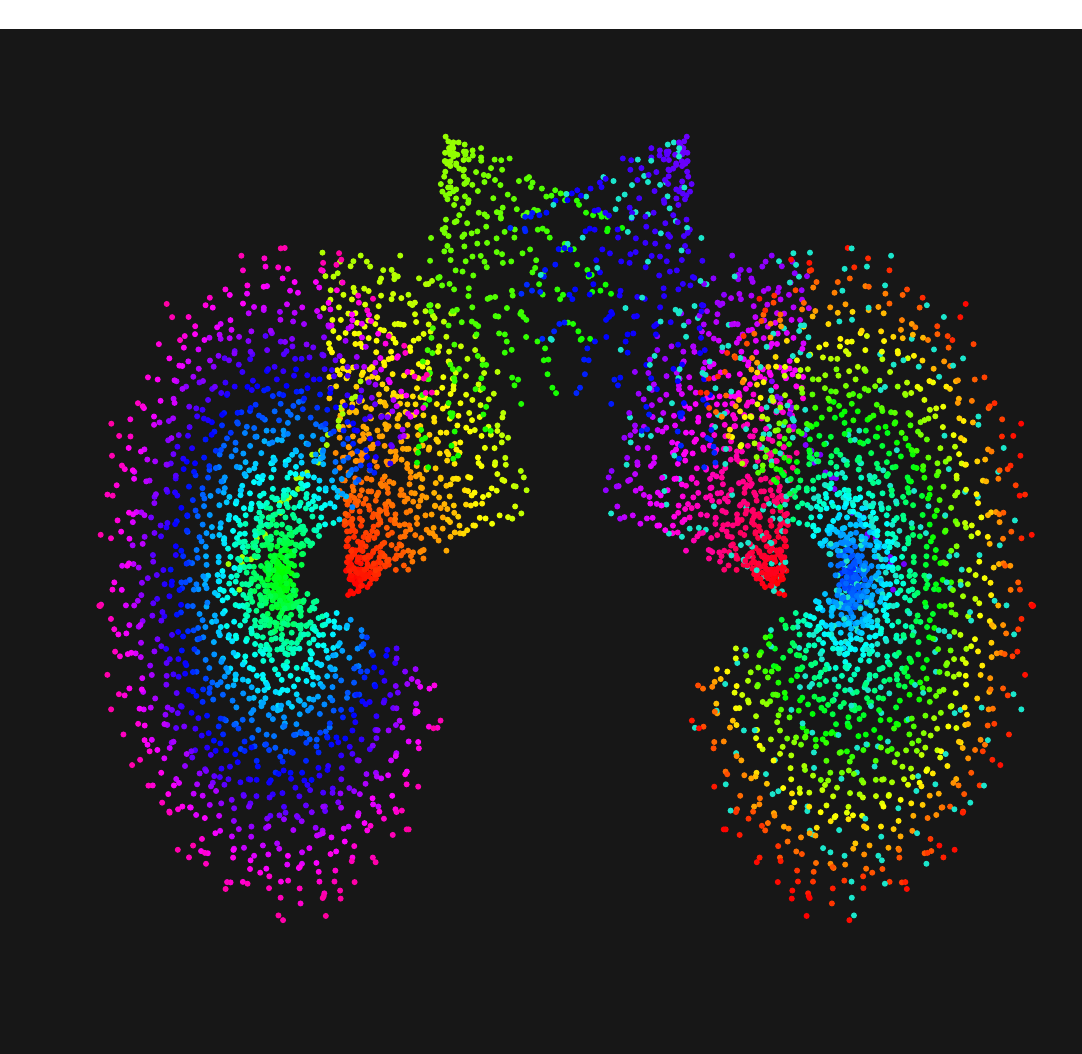}
\includegraphics[trim = 50mm 100mm 50mm 90mm,clip,width=4cm, height=4cm]{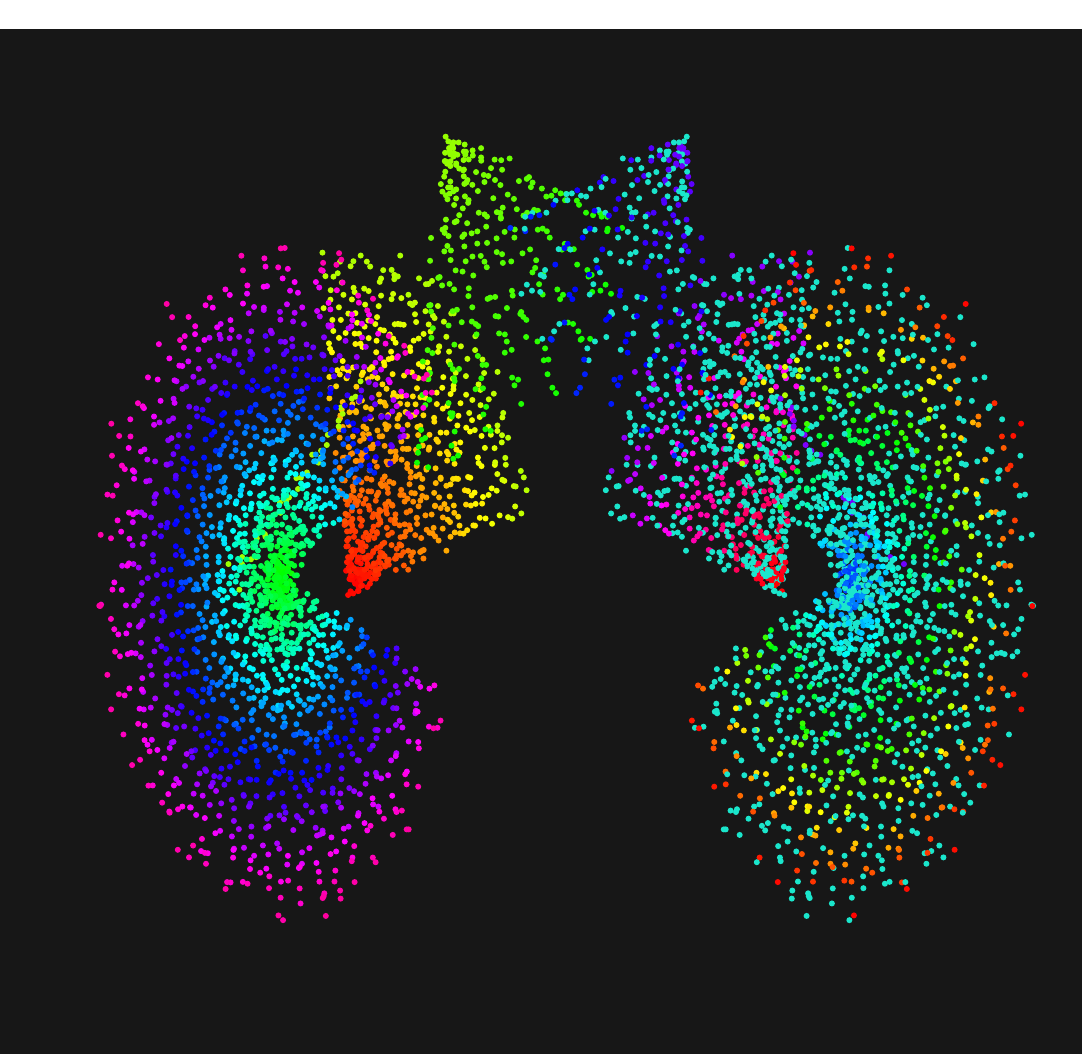}

 \hspace{0.1cm} (a) $\sigma_{break}(c)=0$ \hspace{1.7cm} (b) $\sigma_{break}(c)=0.25$ \hspace{1.25cm} (c) $\sigma_{break}(c)=0.75$

\includegraphics[trim = 50mm 100mm 50mm 90mm,clip,width=4cm, height=4cm]{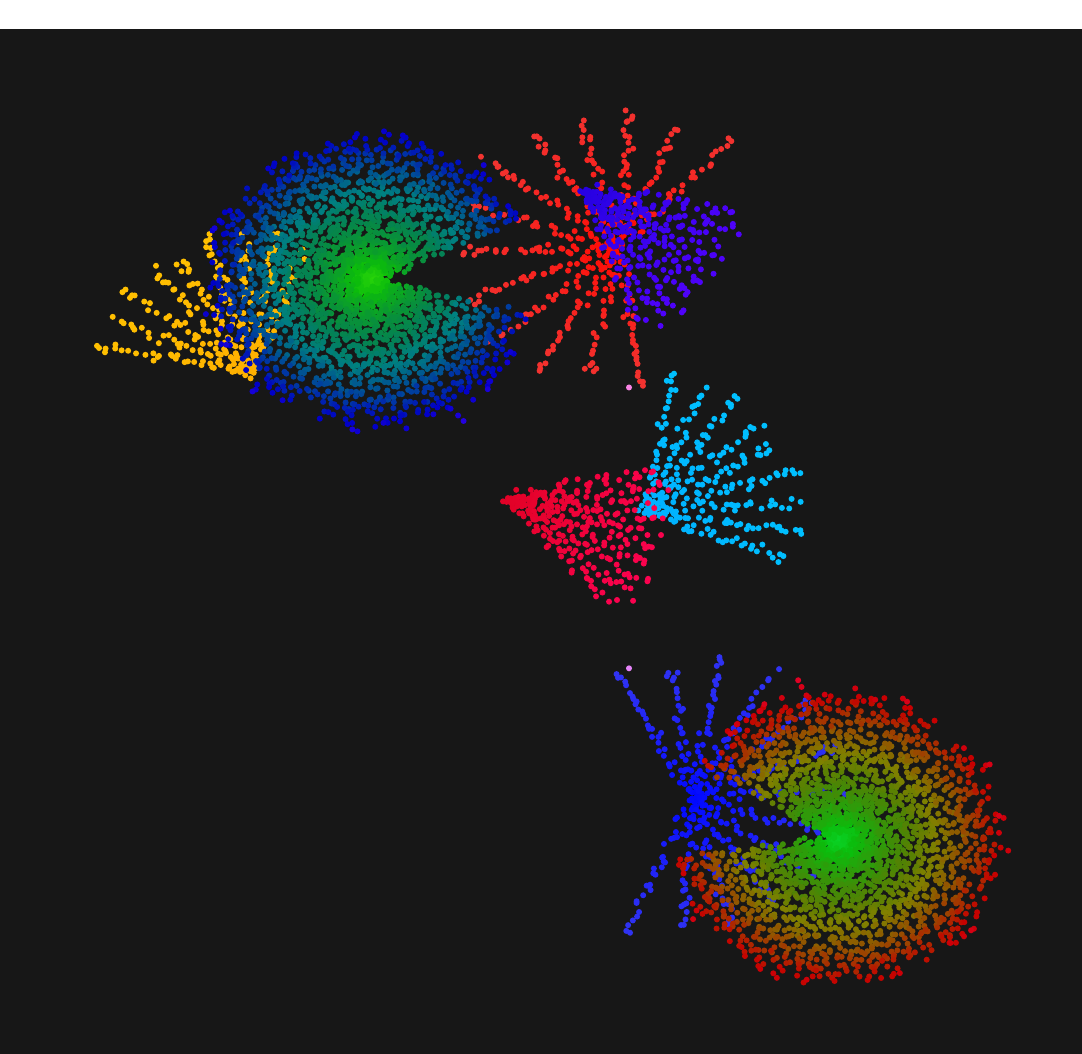}
\includegraphics[trim = 50mm 100mm 50mm 90mm,clip,width=4cm, height=4cm]{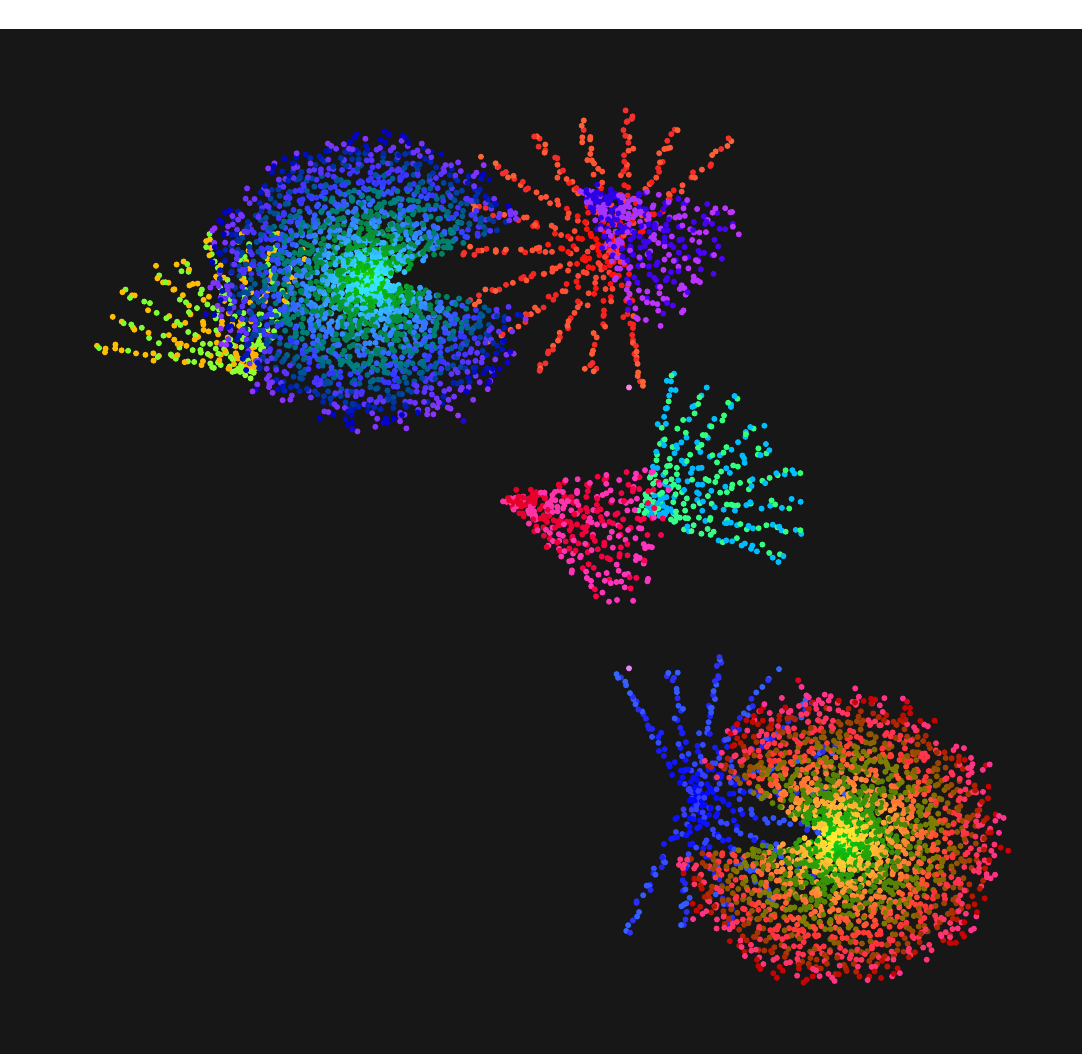}
\includegraphics[trim = 50mm 100mm 50mm 90mm,clip,width=4cm, height=4cm]{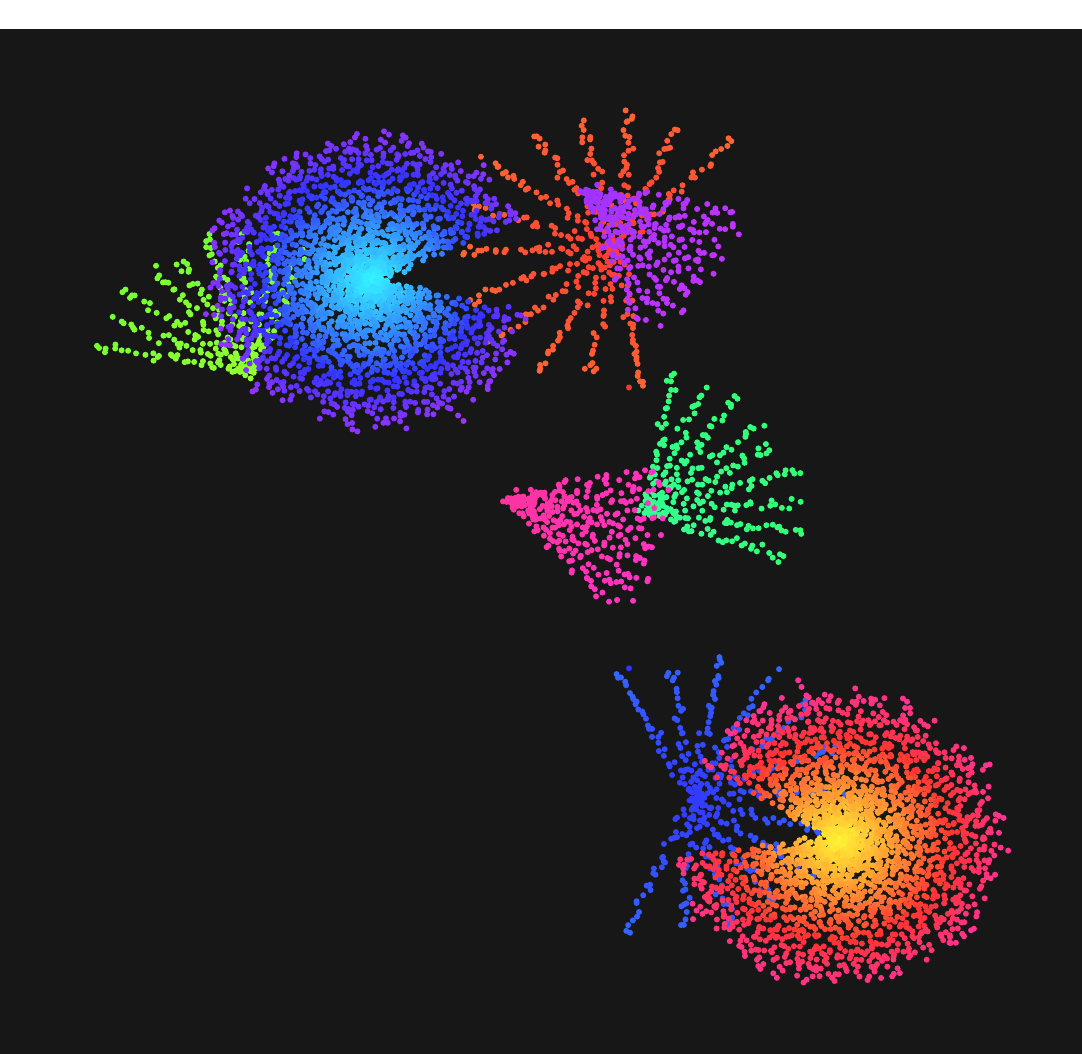}

 \hspace{0.1cm} (d) $\sigma_{break}(c)=0$ \hspace{1.7cm} (e) $\sigma_{break}(c)=0.50$ \hspace{1.25cm} (f) $\sigma_{break}(c)=0.95$

\caption{Images with polychromatic symmetry using a RYB and RGB color wheel   
}
\label{fig:symm3}
\end{figure}

 Note that the images in Fig. \ref{fig:symm1} not only represent pattern symmetry, but also a simple form of color symmetry. This is because in visual art, color symmetry is not about color itself but depends on the context of geometric symmetry. Color symmetry needs the symmetry of a (possibly monochromatic) pattern to induce a permutation of colors. This permutation of colors should be consistent with the geometric symmetry insofar as some (or all) symmetry operations change the colors, while some other  operations (or none) preserve color. For dichromatic images as Fig. \ref{fig:symm1} the color permutation group has degree 2, which is to say there are only 2 colors. We see in the images that the pattern symmetry leads to a color change if the symmetry is by reflection and preserves the color if the symmetry is by translation, which is a simple form of color symmetry.
To obtain a  polychromatic color symmetry we need a permutation group of degree $N$, with $N$ the number of colors involved. Such a color permutation can be considered as a mapping on a color wheel with $N$ slots, see the example of RYB and RGB color wheels with 12 slots in Fig.  \ref{fig:col_wheel}.  A broken color symmetry implies that not all pellets experience the color permutation, but a fraction only. In other words, we perturb the change--or--preserve--color arrangement induced by the color permutation. 

Fig.  \ref{fig:symm2} shows such a color symmetry and also the results of some experiments with color symmetry breaking.
The color permutation is realized using a RYB (red--yellow--blue) color wheel~\cite{itten73,rhyne17}, see also Fig.  \ref{fig:col_wheel}a. 
It is also called the standard artistic color wheel and defines 3 primary colors: \textbf{r}ed, \textbf{y}ellow and \textbf{b}lue. Mixing 2 of these colors each gives the 3 secondary colors: \textbf{o}range (red and yellow), \textbf{p}urple (red and blue) and \textbf{g}reen (yellow and blue). From these 3 primary and 3 secondary colors, another 6 tertiary colors can be derived by mixing: \textbf{ve}rmillion (red and orange), \textbf{am}ber (orange and yellow), \textbf{ch}artreuse (yellow and green), \textbf{te}al (green and blue), \textbf{vi}olet (blue and purple) and \textbf{ma}genta (purple and red).   Fig.  \ref{fig:symm2}a shows an image with full color symmetry. The pattern consists of 3 burrows that each have a symmetric counterpart. Thus, there are 6 burrows in total. The 3 burrows in the lower half of the image are colored with the 3 primary colors according to the RYB color wheel. The symmetric burrows in the upper half of the image are colored with the secondary colors so that the symmetry of the pattern yields the complementary color of the RYB wheel, which is the color exactly opposite on the wheel. Using Cauchy's two--line notation for describing the color permutation group, we can write $\theta=\left(\begin{smallmatrix} \mathbf{r} & \mathbf{y}& \mathbf{b} \\ \mathbf{g} & \mathbf{p} & \mathbf{o}  \end{smallmatrix} \right)$ to express this color symmetry. We now break the color symmetry. Therefore, a fraction of pellets is selected at random with the symmetry breaking rate $\sigma_{break}(c)$ and colored with tertiary colors according to the RYB wheel. 
In some sense, this example of color symmetry breaking finally yields another full symmetry for  $\sigma_{break}(c)=1$, which is  $\theta=\left(\begin{smallmatrix} \mathbf{ch} & \mathbf{te}& \mathbf{vi} \\ \mathbf{ma} & \mathbf{ve} & \mathbf{am}  \end{smallmatrix} \right)$  in two--line notation. The results shown in Fig.  \ref{fig:symm2}b--f   are the intermediate steps between two unbroken symmetries.

The next experiment involves a RGB (red--green--blue) color wheel, which is based on the light model of color and commonly finds  usage in computer graphics~\cite{rhyne17,shev03}, see also Fig. \ref{fig:col_wheel}b. 
Although roughly the RGB color wheel covers the same color spectrum as the RYB color wheel, the colors are distributed differently on the wheel. Particularly, the complementary colors (which are placed opposite on the color wheel) are different. In addition, the warmer colors are spread further around on the RYB wheel, which gives the RGB wheel a somewhat cooler appearance. This gives rise to another color symmetry, see Fig.  \ref{fig:symm3}.  The patterns are obtained by rotation and glide reflection.  Fig. \ref{fig:symm3}a shows a pattern colored by a RGB wheel with broken color symmetry  in Fig. \ref{fig:symm3}b--c. 
For the symmetric pattern in the right half of  the image  fractions of pellets are colored in cyan.
In Fig. \ref{fig:symm3}d--f the complete pattern is mapped in two fractions from a RYB color wheel onto a RGB color wheel.

\subsection{Computational aesthetic measures and analyzing symmetry}

One of the aims of this paper is to analyze how computational aesthetic measures may identify symmetry and symmetry breaking. We use a symmetry measure $\text{SYM}$ that is specifically designed to account for symmetry~\cite{den12,den14,den15}, propose a  refinement of this symmetry measure  and also consider a general computational aesthetic measure,  Benford's law measure $\text{BFL}$~\cite{den14,neu17}, which has shown to scale regularly with design parameters of the sand--bubbler patterns~\cite{rich18}.   The measures are not calculated with the intention to obtain an evaluation of the artistic value, but to have a metric that may reflect design parameters, i.e. the type of symmetry and the symmetry breaking rate.

As symmetry implies that parts of the image in some ways resemble each other, the main algorithmic approach  to symmetry evaluation works by dividing the image and comparing the parts. One method is to define axis along the diagonals of the image (for instance the horizontal, vertical, main and secondary diagonal) and comparing the sections on opposing sides of the axis~\cite{gar17,liu10}.  Another approach is to partition the image into rectangles of equal size and comparing them. This method has been proposed and studied by den Heijer for evolutionary art~\cite{den12,den14,den15}. It proposes to partition the image into 4 areas. We apply this method  here and straightforwardly extended it by considering a finer partitioning into 16 areas.

\begin{figure}[tb]
\center
\includegraphics[trim = 12mm 127mm 130mm 128mm,clip,width=6cm, height=4cm]{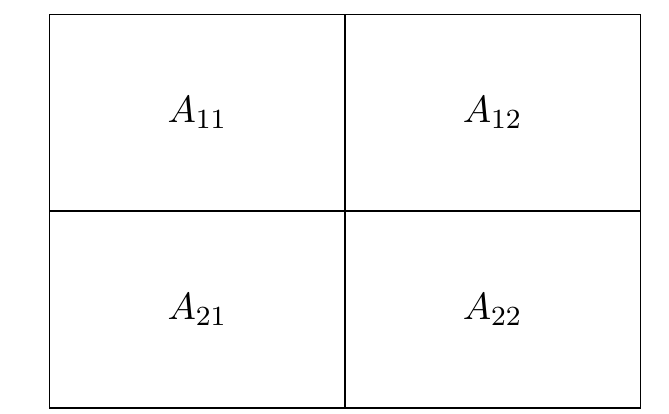}
\includegraphics[trim = 12mm 127mm 130mm 128mm,clip,width=6cm, height=4cm]{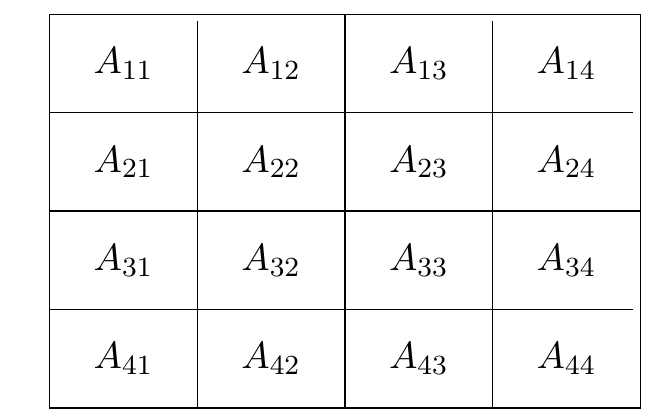}

 \hspace{1cm} (a) \hspace{6cm} (b)

\caption{Calculating of symmetry measures   
}
\label{fig:symm_grid}
\end{figure}

 Given that the image under study is rectangular, the most basic implementation of this method is to divide the image in 4 quadrants, the north--west, north--east, south--west, and south--east corner ($A_{11}$,$A_{12}$,$A_{21}$,$A_{22}$), see Fig. \ref{fig:symm_grid}a. The quadrants are grouped into  4 areas: \textbf{l}eft, \textbf{r}ight, \textbf{t}op and \textbf{b}ottom,  $A_{\mathbf{l}}=A_{11} \cup A_{21}$, $A_{\mathbf{r}}=A_{12} \cup A_{22}$, $A_{\mathbf{t}}=A_{11} \cup A_{12}$, $A_{\mathbf{b}}=A_{21} \cup A_{22}$,  respectively. The implementation as considered by  den Heijer~\cite{den12,den14,den15} also takes into account diagonal similarity. However, numerical experiments (not given in figures) have shown neither  positive nor negative  effects. 
Therefore, and to save computational resources, diagonal similarity is not integrated in the implementation used here. 
 
We compare the  areas by evaluating the differences in intensity for each RGB pixel. In the experiments, we consider the images to have $256 \times 256$ pixels.
The intensity $I_n(i,j)$ of a pixel $(ij)$ belonging to an area $A_{n}$ is obtained as the average of its red ($r$), green ($g$) and blue ($b$) value: $I_{n}(i,j)=(r(i,j)+g(i,j)+b(i,j))/3.$
Thus, the similarity between a pixel $(i,j)$ belonging to an area $A_{n}$ and a pixel $(i,j)$ belonging to an area $A_{m}$ is
\begin{equation}sim(A_{n}(i,j),A_{m}(i,j))= \left\{ \begin{array}{cl} 1 & \text{if} \quad |I_{n}(i,j)-I_{m}(i,j)| < \alpha, \\ 0 & \text{otherwise}
\end{array} \right. \end{equation} with $\alpha$ a difference threshold. In the experiments, there is $\alpha=0.05$ (and  $0 \leq I_{n} \leq 1$).
For similarity of whole areas we average over all pixels:
\begin{equation}sim(A_{n},A_{m})= \frac{1}{\mathcal{I} \mathcal{J}} \sum_{i=1}^{\mathcal{I}} \sum_{j=1}^{\mathcal{J}} sim(A_{n}(i,j),A_{m}(i,j)).
\end{equation}
To define horizontal symmetry of 4 areas $sym_{4h}$, we calculate the similarity between the left and the right area: \begin{equation} sym_{4h}=sim(A_{\mathbf{l}},A_{\mathbf{r}}). \end{equation}
Likewise, vertical  symmetry $sym_{4v}$ is computed by comparing top and bottom:
\begin{equation} sym_{4v}=sim(A_{\mathbf{t}},A_{\mathbf{b}}). \end{equation}
The symmetry measure  $\text{SYM}_4$ taking into account a partition into 4 areas of the image as shown in Fig.  \ref{fig:symm_grid}a is the  average over these two symmetries:
\begin{equation} \text{SYM}_4=(sym_{4h}+sym_{4v})/2. \label{eq:symmeas} \end{equation}
Apparently, the symmetry measure $\text{SYM}_4$ relies upon the assumption that the similarity shows by comparing these 4 areas of the image. A straightforward extension, which is proposed here, compares a finer grid of areas. Therefore, we quarter each of the 4 areas to obtain 16 areas, see Fig.     \ref{fig:symm_grid}b. 
We now define a \textbf{l}eft area by $A_{\mathbf{l}}=A_{11} \cup A_{21} \cup A_{31} \cup A_{41}$, a \textbf{m}iddle \textbf{l}eft by $A_{\mathbf{m} \: \mathbf{l}}=A_{12} \cup A_{22} \cup A_{32} \cup A_{42}$, a \textbf{m}iddle \textbf{r}ight by $A_{\mathbf{m} \: \mathbf{r}}=A_{13} \cup A_{23} \cup A_{33} \cup A_{43}$  and a \textbf{r}ight area by $A_{\mathbf{r}}=A_{14} \cup A_{24} \cup A_{34} \cup A_{44}$. The same is done like--wise for vertical areas: \textbf{t}op, \textbf{m}iddle \textbf{t}op, \textbf{m}iddle \textbf{b}ottom and \textbf{b}ottom.  
For the horizontal symmetry $sym_{16h}$ of 16 areas we now calculate the similarity between the left and the areas to the right \begin{equation} sym_{16h}=(sim(A_{\mathbf{l}},A_{\mathbf{r}})+sim(A_{\mathbf{l}},A_{\mathbf{m} \: \mathbf{l}})+sim(A_{\mathbf{l}},A_{\mathbf{m} \:\mathbf{r}}))/3, \end{equation} while for the vertical symmetry $sym_{16v}$ of 16 areas, we take into account the similarity between the top and the areas below
 \begin{equation} sym_{16v}=(sim(A_{\mathbf{t}},A_{\mathbf{b}})+sim(A_{\mathbf{t}},A_{\mathbf{m} \: \mathbf{t}})+sim(A_{\mathbf{t}},A_{\mathbf{m} \:\mathbf{b}}))/3. \end{equation} Note that by this comparison of areas a bias is imposed towards the left and top of the image. Additional experiments (not depicted in the figures) have shown that the results are qualitatively the same if the bias is towards the right or the bottom.  This appears to be plausible as the images have no natural alignment.  
 
\begin{figure}[tb]
\center
\includegraphics[trim = 2mm 1mm 2mm 0mm,clip,width=6cm, height=4.5cm]{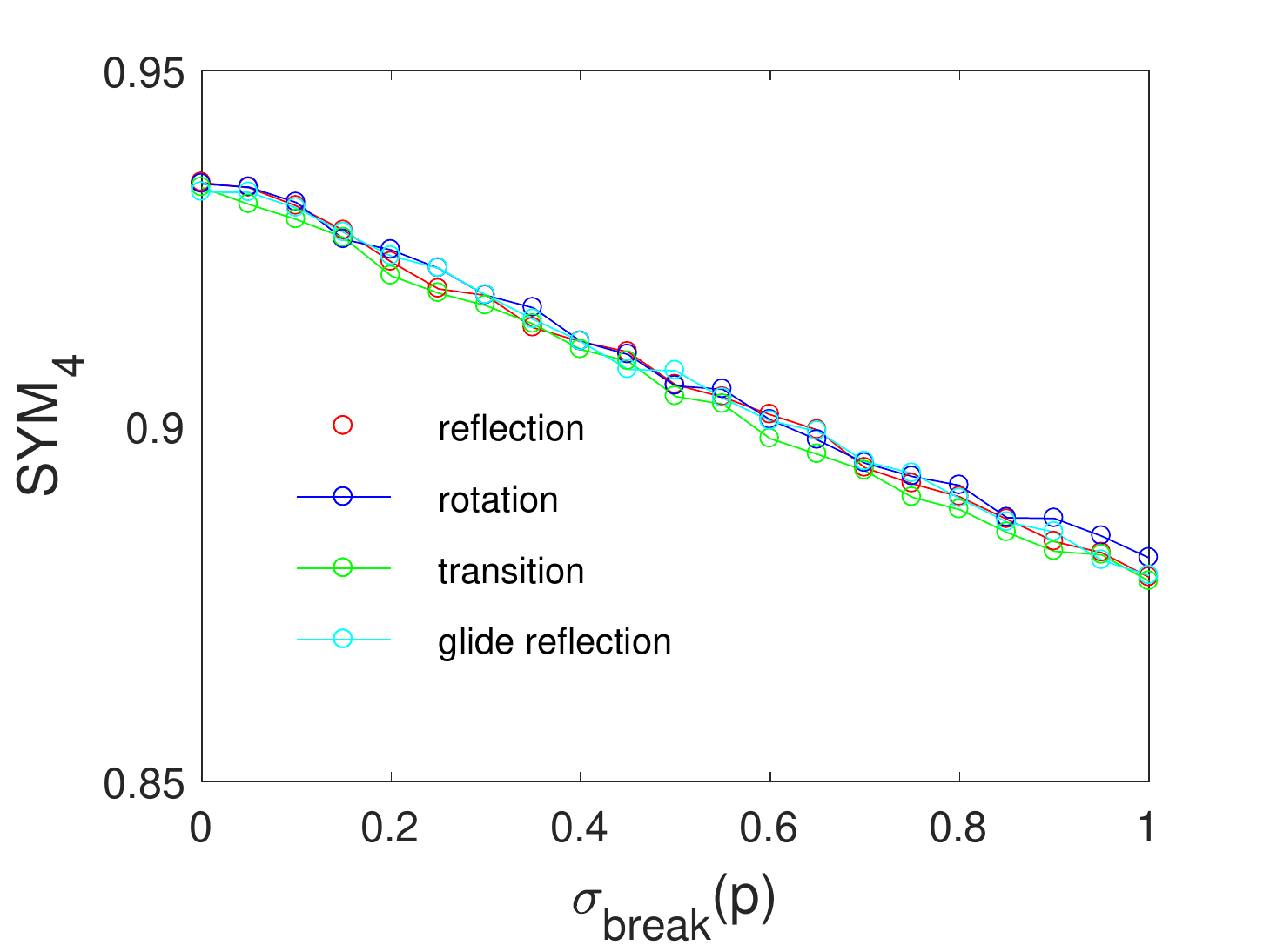}
\includegraphics[trim = 2mm 1mm 2mm 0mm,clip,width=6cm, height=4.5cm]{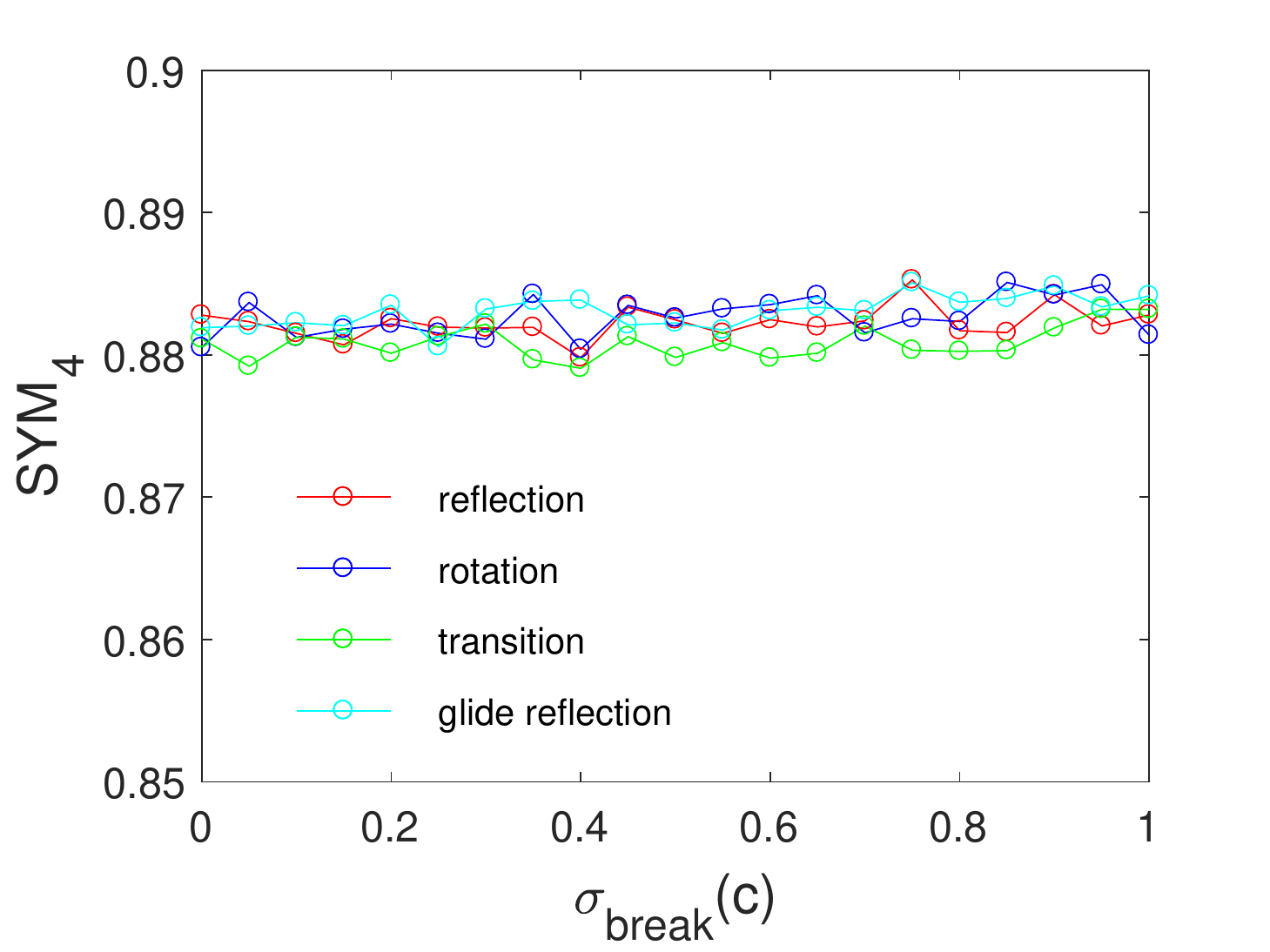}

 \hspace{1cm} (a) \hspace{6cm} (b)
 
\includegraphics[trim = 2mm 1mm 2mm 0mm,clip,width=6cm, height=4.5cm]{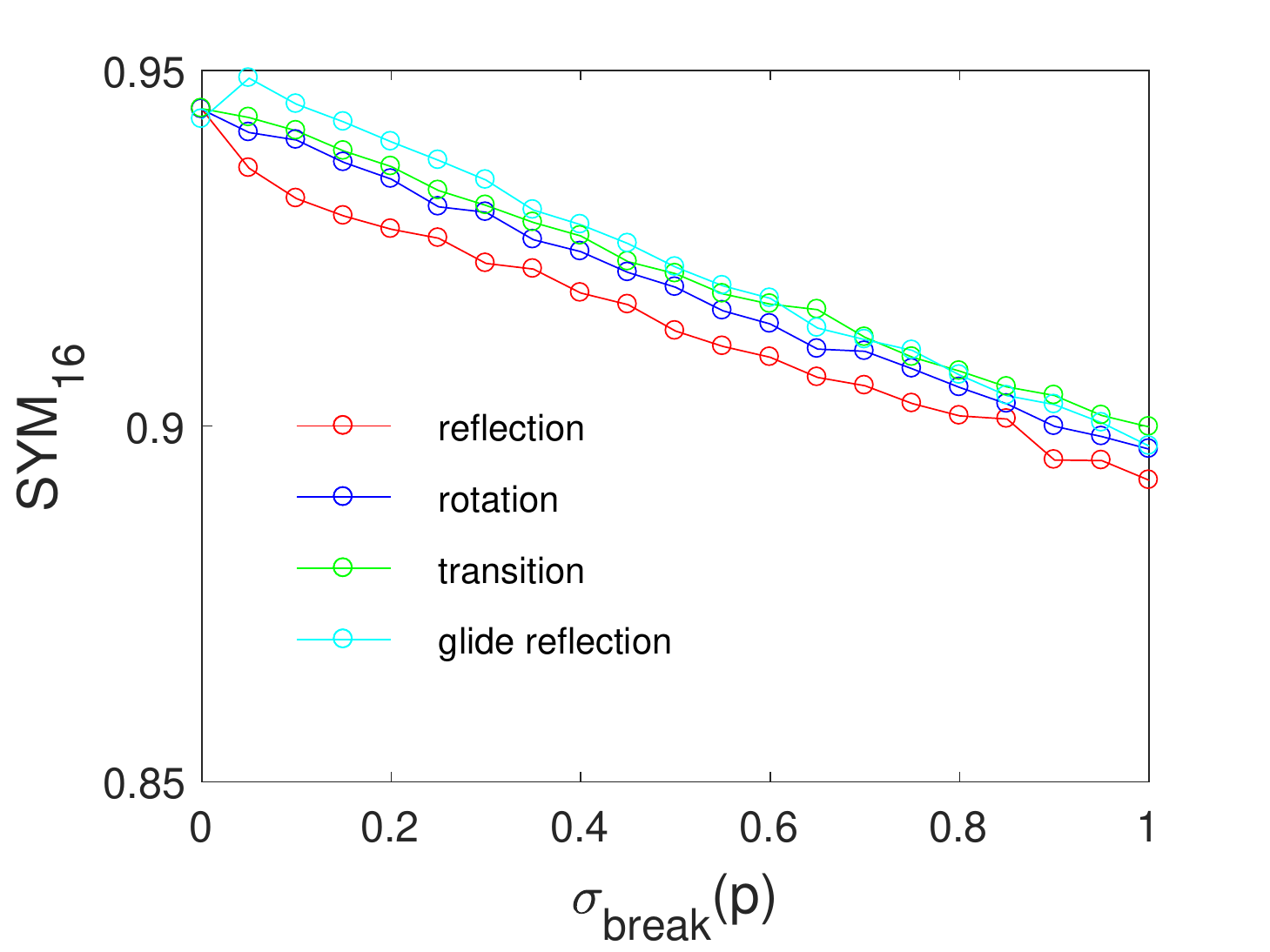}
\includegraphics[trim = 2mm 1mm 2mm 0mm,clip,width=6cm, height=4.5cm]{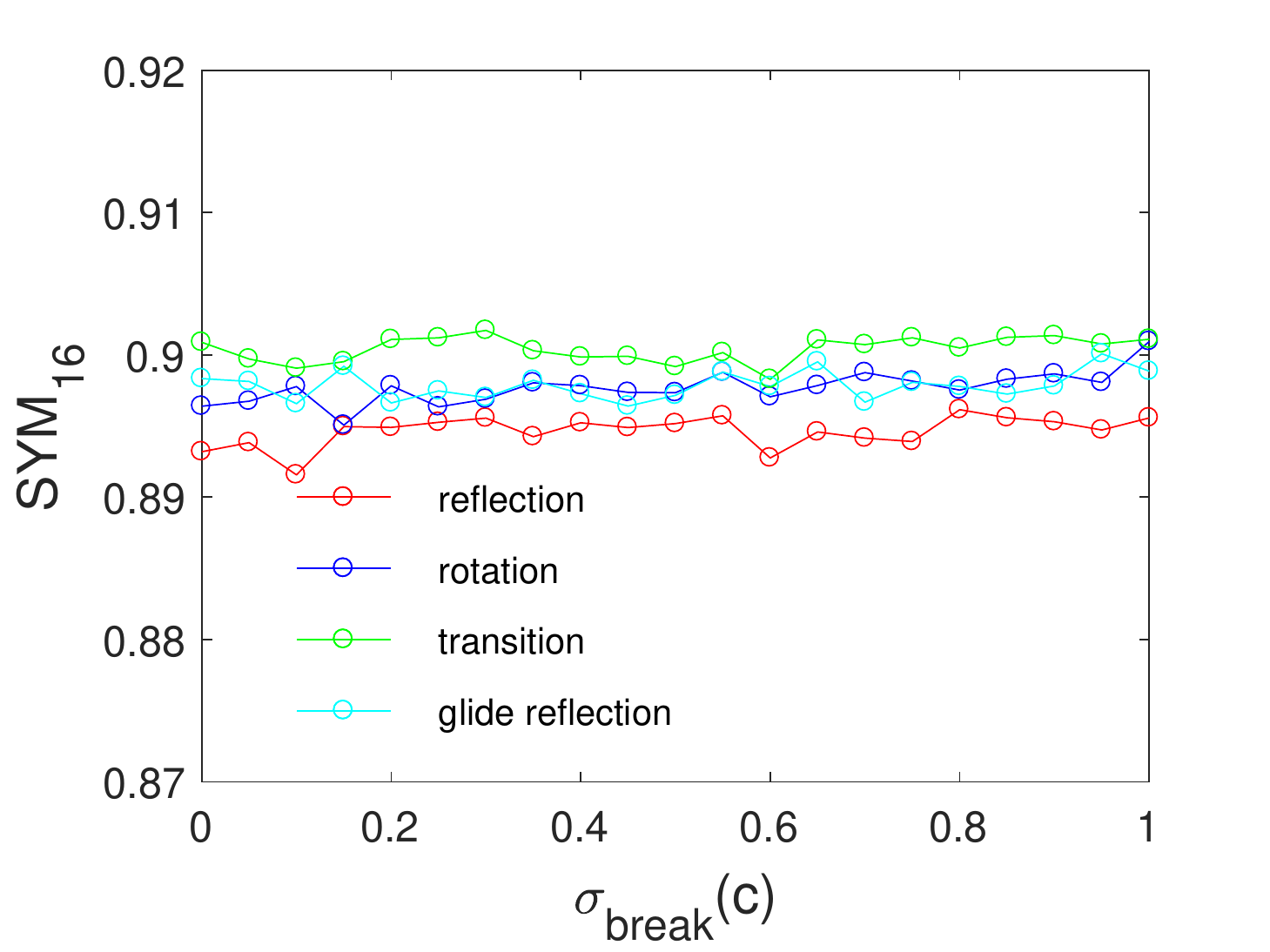}

 \hspace{1cm} (c) \hspace{6cm} (d)

\caption{The symmetry measure $\text{SYM}$, Eq. (\ref{eq:symmeas}), as a function of the symmetry breaking rates $\sigma_{break}$ for both symmetry breaking of pattern and color symmetry breaking   
}
\label{fig:measures}
\end{figure}

The symmetry measure  $\text{SYM}_{16}$ taking into account a partition into 16 areas of the image as shown in Fig.  \ref{fig:symm_grid}b is again the  average over the two symmetries:
\begin{equation} \text{SYM}_{16}=(sym_{16h}+sym_{16v})/2. \label{eq:symmeas16} \end{equation} 
Fig. \ref{fig:measures} shows the results of computational experiments with symmetry and symmetry breaking in digital visual art. The symmetry measure $\text{SYM}_4$, Eq. (\ref{eq:symmeas}), and $\text{SYM}_{16}$, Eq. (\ref{eq:symmeas16}), are shown as a function of the symmetry breaking rates $\sigma_{break}$ for  symmetry breaking of both pattern and color. The results are for 21 equidistant values of the symmetry breaking rate    $\sigma_{break}$  with  $0 \leq \sigma_{break} \leq 1$. 
For each value of   $\sigma_{break}$, 2000 images with patterns were generated according to the algorithmic framework described in~\cite{rich18}. Each image 
is different due to the fact that some design parameters depend on realizations of a random process. Also the color of each pellet is assigned at random. This is done by coloring each pellet with a realization of a random variable uniformly distributed on the RGB color space. Symmetry breaking of the pattern is done by pellets being removed (or made invisible). 
Looking at the results, we see that for pattern symmetry (Fig. \ref{fig:measures}a and \ref{fig:measures}c) the symmetry measures fall almost linearly with the symmetry breaking rate 
 $\sigma_{break}(p)$. For  $\sigma_{break}(p)=0$, where the symmetry is   completely intact, we have the highest values of $\text{SYM}_4$ and $\text{SYM}_{16}$. The values fall for $\sigma_{break}(p)$ getting larger and are smallest for $\sigma_{break}(p)=1$, where symmetry generated by the isometric maps is completely gone. However, we also see that the values of the symmetry measures for  $\sigma_{break}(p)=1$ are not really small. We find   $\text{SYM}_4 \approx 0.88$ and  $\text{SYM}_{16} \approx 0.9$.  
This can be explained by even a single sand--bubbler pattern displaying a considerable degree of symmetry. Look, for instance, at the pattern on the left--hand side of Fig. \ref{fig:symm1}a. Here, the trenches of the pattern form hands around the center point for more than a semicircle. Thus, there is symmetry in itself, and the symmetry measures  $\text{SYM}_4$ and $\text{SYM}_{16}$ account for it. However, generating additional symmetry by the isometric maps   (\ref{eq:reflect})--(\ref{eq:glide})  increases 
the value of the symmetry measures even more, which shows their ability to identify different shades of symmetry. 
The main difference between the curves for  $\text{SYM}_4$ and $\text{SYM}_{16}$ is that for the former there is no distinguishable difference between the different types of symmetry. For  $\text{SYM}_{16}$ such a differentiating is partly possible, with reflection (\ref{eq:reflect}) giving slightly smaller values than rotation (\ref{eq:rotate}), transition (\ref{eq:transl}) and glide reflection (\ref{eq:glide}).

The curves for color symmetry breaking are shown with Fig.   \ref{fig:measures}b and \ref{fig:measures}d. Symmetry breaking of colors is done by shifting the colors of the selected pellets randomly through the RGB color space. The results show that a higher degree of symmetry breaking (that is, a higher percentage of pellets that change their color) does not give very different values of the symmetry measures. In fact, both  $\text{SYM}_4$ and $\text{SYM}_{16}$ slightly drift but have essentially the same value for all $0 \leq \sigma_{break}(c) \leq 1$. This is a consequence of how the symmetry measures are calculated.  For a given collection of pellets they account for difference in the spatial distribution of their average RGB values. Thus, a random shift through the RGB color space does not systematically alter intensity. Such a alteration can be achieved if  breaking the color symmetry has a bias towards one of the RGB components. Experiments indeed show this to be the case.   However, such a bias is completely arbitrary and does not make the symmetry measure truly able to quantify color symmetry breaking. Finally, it can be observed that for color symmetry breaking (as for pattern symmetry breaking)  $\text{SYM}_{16}$ allows to differentiate between types of symmetry, while  $\text{SYM}_4$  does not. 

\begin{figure}[tb]
\center
\includegraphics[trim = 2mm 1mm 2mm 0mm,clip,width=6cm, height=4.5cm]{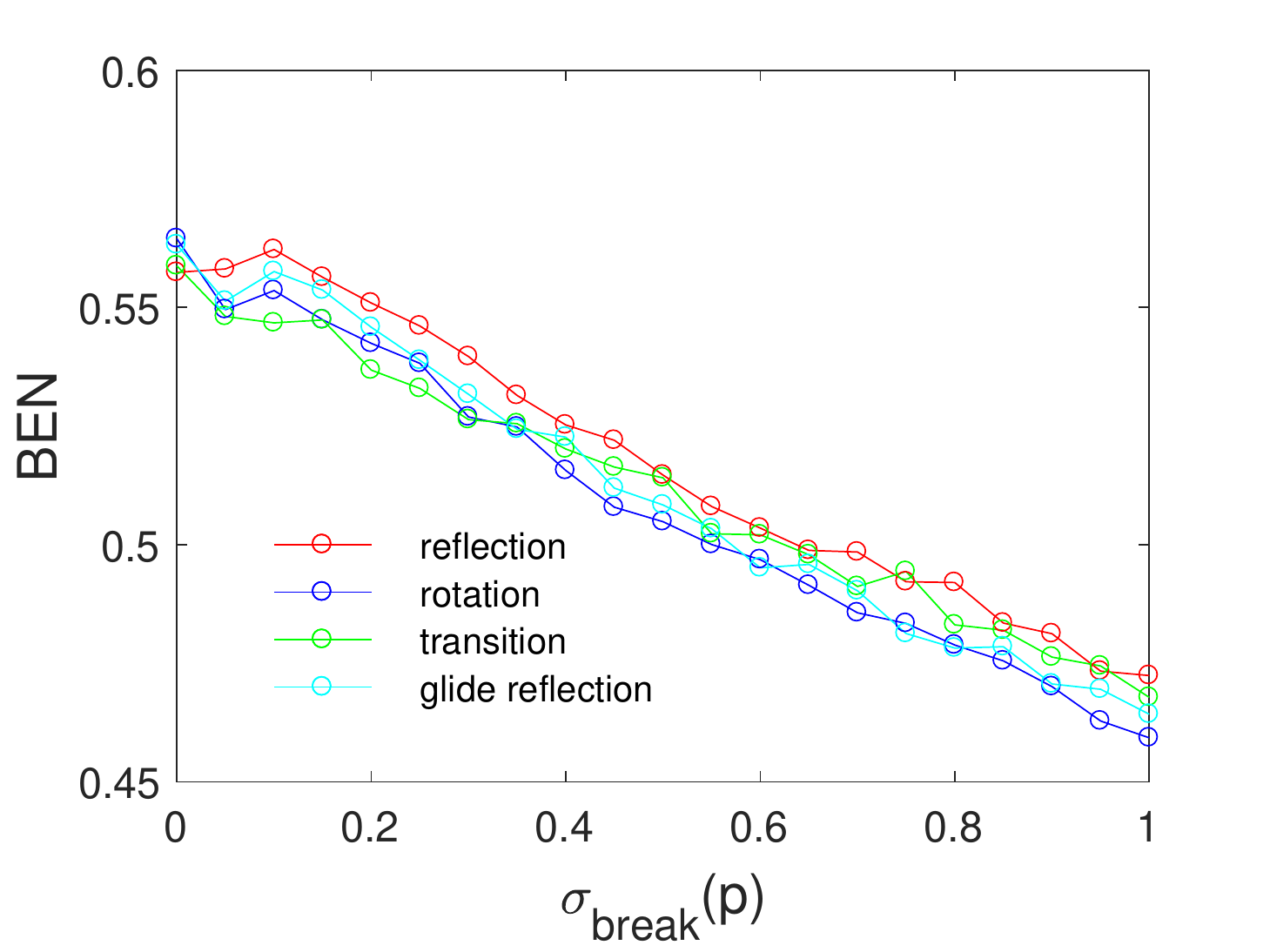}
\includegraphics[trim = 2mm 1mm 2mm 0mm,clip,width=6cm, height=4.5cm]{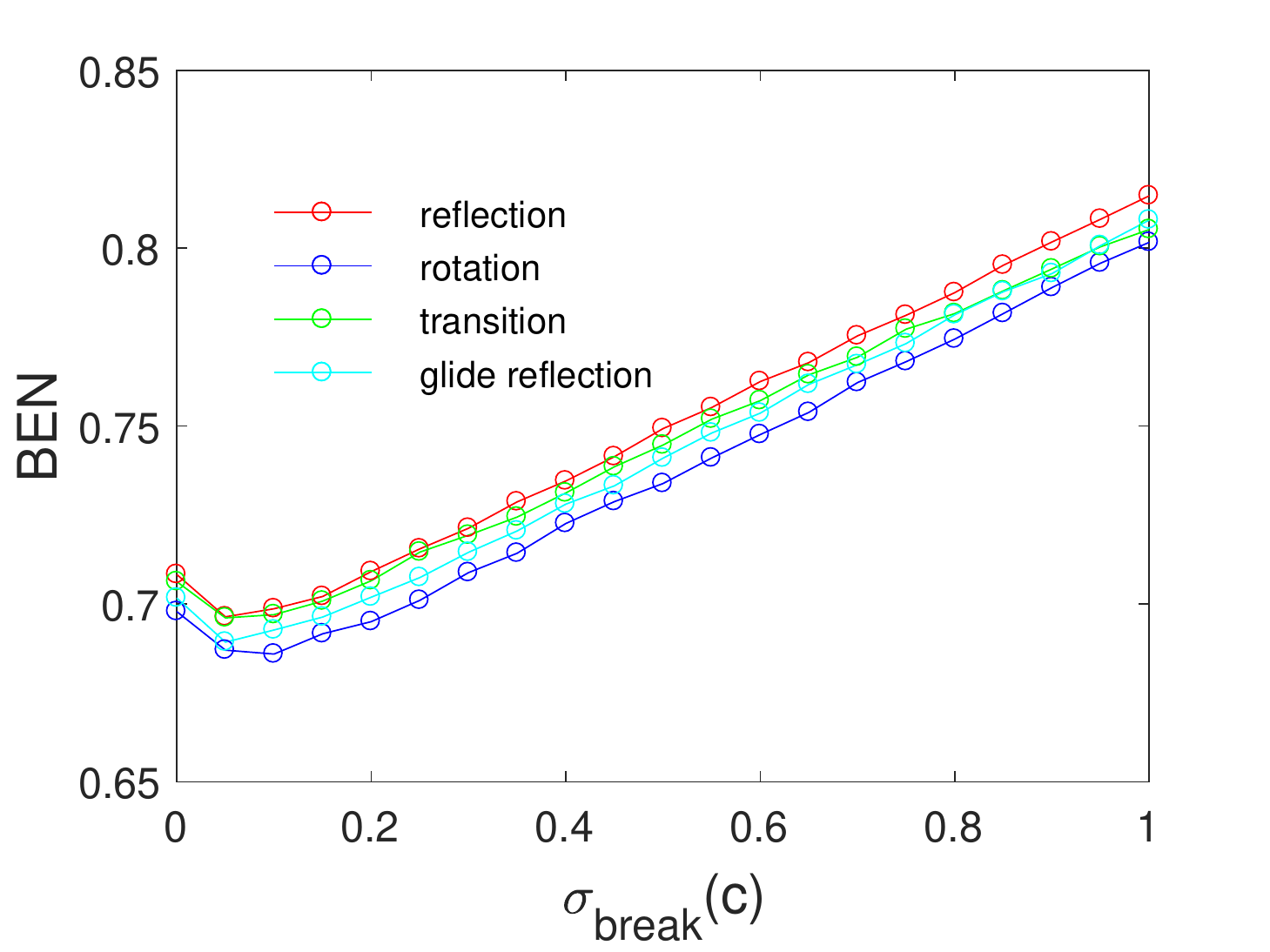}

\caption{The Benford's law measure $\text{BFL}$, Eq. (\ref{eq:bfl}),  shown as a function of the symmetry breaking rates $\sigma_{break}$ for both symmetry breaking of pattern and color symmetry breaking   
}
\label{fig:measures_ben}
\end{figure}

It may be interesting to note that an even finer grid of areas does not improve results. Additional experiments with 64 areas (not shown in figures due to brevity) have shown 
that the curves for the isometric maps (\ref{eq:reflect})--(\ref{eq:glide}) lump together as for 4 areas. This immediately suggests the conjecture that there is a grid optimal for differentiating between types of symmetry, which may depend on the granularity of the pellets. This appears understandable as the pellets have a finite size, which is relative to the size of the image. The pellets have no structure on every scale as for instance have fractals, which is the main reason for not using fractal dimensions to evaluate the images.

 The Benford's law measure $\text{BFL}$~\cite{den14,neu17}  is a measure of the naturalness of an image. It is calculated by taking the distribution of  the luminosity of an image and comparing it to the Benford's law distribution:
\begin{equation} \text{BFL}= 1- d_{total}/d_{max} \label{eq:bfl}\end{equation}
where  $d_{total}= \sum_{i=1}^9 \left(H_{I}(i)-H_{B}(i) \right)$ and  $d_{max}=1.398$ is the maximum possible value of $d_{total}$. Furthermore,  
\begin{equation} H_B=\left( 0.301,0.176,0.125,0.097,0.079,0.067,0.058,0.051,0.046 \right) \end{equation}  is the Benford's law distribution  and
 $H_{I}$ is the normalized sorted $9$--bin histogram of the luminosity of the image. The luminosity  of the image is calculated from the luminosity $L(i,j)$ of each pixel $(ij)$ of the image by taking the weighted sum of the red ($r$), green ($g$) and blue ($b$) values: 
$L(i,j)=0.2126 \cdot r(i,j)+0.7152 \cdot g(i,j)+0.0722\cdot b(i,j).$ 
The results are shown in Fig. \ref{fig:measures_ben}. We see that the Benford's law measure $\text{BFL}$ decreases with rising symmetry break rate $\sigma_{break}(p)$ and increases 
with $\sigma_{break}(c)$. Also, the different types of symmetry can be distinguished. Here, rotation gives the lowest values and reflection the highest. Comparing the two computational measures with respect to their abilities for identifying symmetry, we may conclude that   Benford's law measure is as good as the symmetry measure $\text{SYM}_{16}$ and even more informative for color symmetry. 
This may seem counter--intuitive as the  $\text{BFL}$ evaluates the whole image, while the symmetry measures $\text{SYM}_4$ and $\text{SYM}_{16}$ compare subsections. Presumable, luminosity in connection with a comparison to  the  Benford's law distribution works better than comparing subsections on the basis of intensity differences. Further work is needed to clarify these relationships and also to study if other quantifiers  of pixels than
 intensity and luminosity are an even more suitable option.

\section{Conclusions and future work}

In more abstract terms symmetry means ``immunity to a possible change''~\cite{ros95}. There is the possibility of change and something remains unchanged if the change actually occurs. Such an understanding, however, also suggests the interpretation that symmetry implies redundancy. If a change does not actually change  something, we may learn nothing new.
Thus, to induce some kind of novelty weaker forms of symmetry are desirable as well. 
Between complete symmetry in a mathematical sense and no symmetry at all, there may be intermediate states, which can be seen as different degrees of symmetry breaking. 
In the light of these ideas, this paper discusses creating and analyzing symmetry and broken symmetry in digital art. Its focus  is not so much on computer--generating artistic images, but rather on concepts, templates and tools for incorporating symmetry and symmetry breaking  into the creation process.  Using the example of sand--bubbler patterns as a starting point, all four types of isometric symmetry in two--dimensional  space are employed. In addition, also color symmetry is considered  and realized as a color permutation consistent with the isometries. Generating images  and  an analysis by computational aesthetic measures is also addressed. The main result is that the Benford's law measure and to some extend also  den Heijer's symmetry measures are able to identify different types of symmetry and symmetry breaking in images. 

The visual and analytic results presented in the paper only cover a small subset of possible designs.  Thus, future work could focus on exploring the design space to a greater extent.
For instance, the color permutations discussed in connection with color symmetry 
only considered hue, but could also include saturation or  lightness. 
Furthermore, whole patterns could be used as a motif or building block to create more complex pattern, for instance by 
combining or repeating isometries. Thus, by employing the concept of wallpaper or frieze groups~\cite{cox86,thom12} images could be produced that broaden the spatial scope discussed so far in this paper.

\end{document}